\documentclass[10pt,twocolumn,letterpaper]{article}
\usepackage{iftex}
\ifPDFTeX
\fi
\usepackage{fix-cm}
\usepackage[pagenumbers]{cvpr}
\usepackage[T1]{fontenc}
\usepackage[utf8]{inputenc}
\usepackage{graphicx}
\usepackage{amsmath,amssymb}
\usepackage{booktabs,multirow,array,tabularx}
\usepackage{caption}
\usepackage{stfloats}
\usepackage{afterpage}
\usepackage{balance}
\usepackage{xcolor}
\usepackage{microtype}
\newcommand{\compacttable}{%
  \renewcommand{\arraystretch}{0.88}%
  \setlength{\aboverulesep}{0.25ex}%
  \setlength{\belowrulesep}{0.25ex}%
  \setlength{\extrarowheight}{0pt}}
\newcommand{\method}{WorldReforge}
\newcommand{\dataset}{Reforge}
\newcommand{\NA}{\textemdash}
\newcommand{\para}[1]{\noindent\textbf{#1.}\ }
\usepackage{placeins,enumitem,listings}
\newcommand{\tablefont}{\small\renewcommand{\arraystretch}{1.08}}
\newcolumntype{Y}{>{\raggedright\arraybackslash}X}
\setlist[enumerate]{leftmargin=*,itemsep=2pt,topsep=3pt}
\definecolor{linkblue}{rgb}{0.21,0.49,0.74}
\usepackage[breaklinks,colorlinks,allcolors=linkblue,hyperfootnotes=false]{hyperref}
\usepackage[capitalize]{cleveref}

\date{}
\makeatletter
\renewcommand{\@maketitle}{%
  \newpage
  \null
  \vskip 8pt
  \begin{center}
    {\Large\bfseries\@title\par}
    \vskip 12pt
    {\large\lineskip .5em
      \begin{tabular}[t]{@{}c@{}}\@author\end{tabular}\par}
    \vskip 10pt
  \end{center}%
}
\makeatother

\newcommand{\papertitle}{Beyond the Happy Path:\\Generalizing Robot World Models from Simulated Failures}
\newcommand{\papertitleplain}{Beyond the Happy Path: Generalizing Robot World Models from Simulated Failures}
\newcommand{\paperauthorsplain}{Zhiyue Zhao, Mingyu Liu, Bolin Zhang, Liyang Li, Guangyu Huang, Canyu Zhao, Jin-Chuan Shi, Haokai Zhang, Hao Chen, Chunhua Shen}
\author{%
\parbox{\textwidth}{%
  \centering
  \mbox{Zhiyue Zhao\textsuperscript{1,2,*}}\quad
  \mbox{Mingyu Liu\textsuperscript{1,2,*}}\quad
  \mbox{Bolin Zhang\textsuperscript{1}}\quad
  \mbox{Liyang Li\textsuperscript{1}}\quad
  \mbox{Guangyu Huang\textsuperscript{1}}\quad
  \mbox{Canyu Zhao\textsuperscript{1}}\quad
  \mbox{Jinchuan Shi\textsuperscript{1}}\quad
  \mbox{Haokai Zhang\textsuperscript{1}}\quad
  \mbox{Hao Chen\textsuperscript{1,\textdagger}}\quad
  \mbox{Chunhua Shen\textsuperscript{1,\textdagger}}
  \\[7pt]
  \mbox{\textsuperscript{1}Zhejiang University}\qquad
  \mbox{\textsuperscript{2}Shanghai Innovation Institute}
}%
}

\newcommand{\authorfootnotes}{%
  \begingroup
    \renewcommand{\thefootnote}{\fnsymbol{footnote}}%
    \footnotetext[1]{Equal contribution.}%
    \footnotetext[2]{Corresponding authors.}%
  \endgroup
}

\hypersetup{%
  pdftitle={\papertitleplain},
  pdfauthor={\paperauthorsplain},
  pdfsubject={WorldReforge preprint}%
}
\title{\papertitle}

\begin{document}
\makeatletter
\twocolumn[{%
  \@maketitle
  \centering
  \includegraphics[width=\textwidth]{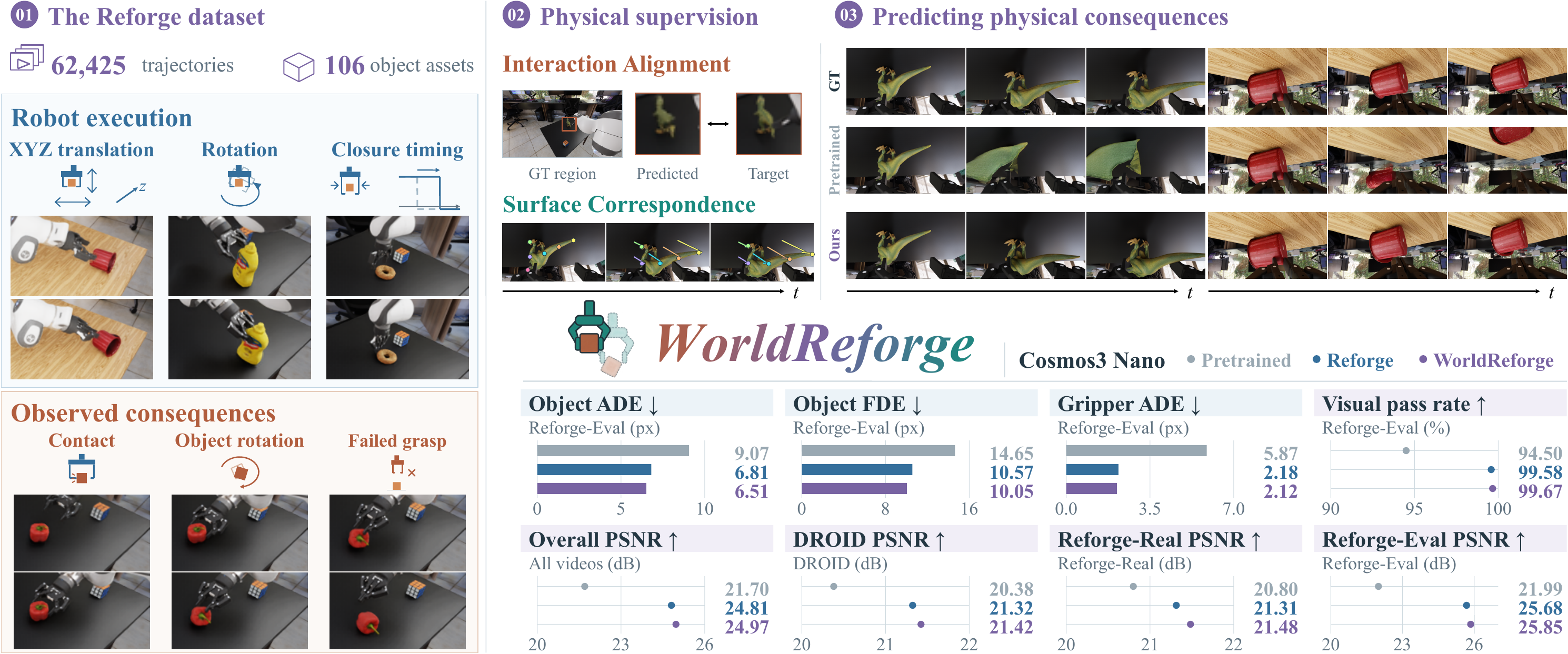}
  \captionsetup{hypcap=false}
  \captionof{figure}{\method{} combines structured interaction experience with physical supervision. 01 The Reforge dataset links execution variations to observed consequences. 02 Interaction Alignment emphasizes interaction regions, while Surface Correspondence supervises surface motion across frames. 03 Consecutive predictions and Cosmos3 Nano comparisons show more accurate object and gripper motion and improved reconstruction in simulated and real videos.}
  \label{fig:teaser}
  \vspace{\dbltextfloatsep}
}]
\makeatother
\authorfootnotes

\begin{abstract}
Robot world models are increasingly used to evaluate and plan policies, yet robot training data are dominated by successful demonstrations. Off the happy path, when a grasp is misaligned or mistimed and the object is pushed, tilted, or slips, their predictions break down. Across five pretrained action-conditioned models, object-motion error is at most 10.4\% lower than that of a static baseline that copies the first frame, and two models are worse than this baseline. We present \textbf{\method{}}, which teaches world models what happens when manipulation goes wrong. We construct the \textbf{\dataset{}} dataset of 62k simulated trajectories by perturbing reference grasps in position, orientation, and closure timing, recording each resulting push, slip, or failed pickup with ground-truth object pose, contact, and surface points. \method{} turns this ground truth into two training signals without additional inference cost. \textbf{Interaction Alignment} focuses reconstruction on the small object and contact regions that a global loss underweights, and \textbf{Surface Correspondence} trains backbone features to follow the same physical surface points across frames. On Cosmos3 Nano, \method{} reduces object trajectory error by 28.2\% overall and by 35.0\% on unseen objects, and it lowers the error in predicting how two actions' outcomes diverge by 43.3\%. Although trained only in simulation, it also improves reconstruction on DROID and on real failure videos. Controlled data studies show that failed contact and successful holding teach complementary physics: failed contact teaches how objects respond, and holding teaches how a grasped state persists.
Our code, data, and models are available at \url{aim-uofa.github.io/WorldReforge/}.
\end{abstract}

\section{Introduction}
\label{sec:intro}
Action-conditioned world models increasingly replace real-world rollouts for policy evaluation, data generation, and planning~\cite{worldarena,worldoracle,ewmbench,dreamdojo}. Their value depends on predicting how different actions produce different outcomes. Manipulation is where this requirement is hardest to meet. A grasp shifted by a centimeter or closed too early does not simply fail. The object is pushed, tilted, or slips from the fingers. Yet their robot training data are dominated by successful demonstrations~\cite{openx,droid}, so this branch of physics is rarely observed.

We find that this happy-path bias is severe. On our simulated evaluation, five pretrained models from three families~\cite{cosmos3,dreamdojo,lingbot} achieve object-motion error at most 10.4\% lower than a static baseline that copies the first frame, and two are worse (\cref{tab:main}). Their predictions keep objects near the initial pose even when the recorded object is pushed or rotated (\cref{fig:qualitative}). Such predictions risk overrating the policies being evaluated~\cite{curewm,fact}.

Recent work fine-tunes on failures obtained by perturbing or executing actions~\cite{curewm,hfwm,fact,dreamtrue}. These methods keep the original generative objective~\cite{curewm}, supervise outcomes through progress or reward signals~\cite{fact,hfwm}, or learn failure dynamics without ground-truth failure videos~\cite{hfwm}. Two limitations remain. First, contact dynamics occupy a small fraction of each frame,
so a global reconstruction loss gives them little weight. Second, outcome-level signals indicate whether an interaction failed, not how the object moved. Simulation provides what real robot logs lack: failures can be produced at controlled offsets and timings, safely and at scale, each with exact object pose, contact, and surface geometry.

We present \textbf{\method{}}, a post-training framework that exploits this ground truth (\cref{fig:teaser}). The \textbf{\dataset{}} dataset contains 62k simulated trajectories over 96 objects, generated by perturbing successful reference grasps in translation, rotation, and closure timing. Trajectories link execution variations to observed outcomes, including contact formation and loss, pushing, tilting, slipping, and whether a grasp is established and maintained. These labels support controlled training compositions. Reforge-Real provides 300 real failure trajectories for evaluation. \method{} adds two objectives to the generative loss. \textbf{Interaction Alignment} (IA) reconstructs object and contact regions with a dedicated, area-normalized signal. \textbf{Surface Correspondence} (SC) requires intermediate backbone features to locate the same physical surface points across frames, supervised by simulator-projected trajectories. Neither adds inference-time parameters.

Post-training on \dataset{} alone reduces object error by about a quarter for Cosmos3 Nano and LingBot-VA, and IA and SC further lower or match every motion error across all three model families. On Cosmos3 Nano, \method{} reduces object ADE by 28.2\% overall and by 35.0\% on objects unseen during post-training, and it predicts the divergence between two actions' outcomes with 43.3\% lower error. Trained only in simulation, it improves PSNR by 1.04\,dB on DROID and 0.68\,dB on real failure videos. Controlled data studies reveal a division of labor. Adding failures in which the gripper closes without securing the object reduces object error on such contacts by 18.0\%, while successful holding remains the best teacher for keeping a grasped object stable.

Our contributions are:
\begin{list}{$\cdot$}{\setlength{\leftmargin}{1em}\setlength{\labelwidth}{.5em}\setlength{\labelsep}{.5em}\setlength{\topsep}{2pt}\setlength{\itemsep}{2pt}\setlength{\parsep}{0pt}\setlength{\partopsep}{0pt}\interlinepenalty=10000}
\item A diagnosis that pretrained action-conditioned world models barely outperform a static predictor under failed contact, and the \dataset{} dataset, combining 62k simulated trajectories with dense physical ground truth, 300 real failure trajectories, and a paired-action protocol for measuring response differences.
\item \method{}, which converts simulator ground truth into interaction-region reconstruction and cross-frame surface correspondence objectives without inference-time cost.
\item Evidence that simulated failures improve object and gripper prediction across model families, transfer to unseen objects and real-world video, and play a role complementary to successful experience.
\end{list}

\section{Related Work}
\label{sec:related}
\subsection{Robot World Models}
Early action-conditioned video predictors model pixel motion~\cite{finn2016}; Visual Foresight uses predictions for control~\cite{visualforesight}, and SV2P models uncertain futures~\cite{sv2p}. World Models learns compressed dynamics~\cite{worldmodels,liu2026stamo,liu2026perfect,yang2025learning}, while PlaNet and DreamerV3 use latent rollouts for planning and policy learning~\cite{planet,dreamerv3,su2026world}. Genie learns controllable latent actions from unlabeled video~\cite{genie}, RoboDreamer composes predictions from language primitives~\cite{robodreamer}, and Diffusion Forcing combines causal generation with per-token noise levels~\cite{diffusionforcing}. UniSim learns interactive visual simulation~\cite{unisim}; Cosmos3 integrates multimodal generation and action tasks~\cite{cosmos3,cosmosmatrix}; DreamDojo transfers knowledge from human videos~\cite{dreamdojo}; and LingBot-VA jointly models video and actions~\cite{lingbot}. IRASim uses frame-level action conditioning~\cite{irasim}, and Ctrl-World combines multi-view generation with pose-conditioned memory~\cite{ctrlworld}. WorldReforge combines successful and failed experience for contact-driven object prediction.

\subsection{Post-training for Physical Alignment}
Real-world interaction data and demonstration-based learning support robot manipulation~\cite{openx,advdex,bridgev2,robomimic,wang2025vq,liu2026perturbot,liu2025gae}. RoboCasa and robosuite support simulated manipulation~\cite{robosuite,robocasa}, while Kubric provides synthetic videos with dense geometric annotations~\cite{kubric}. AHA constructs failure examples for detection and reasoning~\cite{aha}. FACT and CureWM learn from executed failures~\cite{fact,curewm}; DreamTrue and HF-WM use counterfactual or reward-guided training~\cite{dreamtrue,hfwm}; WorldSync aligns generation with actions~\cite{worldsync}. Feature alignment complements training: REPA aligns denoising features with pretrained visual representations~\cite{repa}, VideoREPA aligns spatiotemporal token relations~\cite{videorepa}, and PhysisForcing aligns feature trajectories and semantic relations~\cite{physisforcing}. ABot-PhysWorld uses decoupled discriminative feedback for preference optimization~\cite{abotphysworld}. WorldReforge uses physical ground truth for local reconstruction and surface correspondence.

\subsection{Evaluating Video and Interaction}
Physion evaluates physical prediction in controlled scenarios~\cite{physion}; WorldOracle compares predictions with real executions~\cite{worldoracle}; WorldFailBench evaluates failure causes, stages, and boundaries~\cite{worldfailbench}. Forecasting research motivates careful metric comparisons~\cite{hyndman2006}, and TAP-Vid provides point-tracking protocols~\cite{tapvid}. RAFT estimates optical flow~\cite{raft}; PIPs tracks individual points through occlusion~\cite{pips}; TAPIR refines point trajectories~\cite{tapir}; and CoTracker jointly models tracks, with CoTracker3 scaling supervision through pseudo-labeling~\cite{cotracker,cotracker3,sam2tap}. VBench and VideoScore assess video quality across multiple dimensions~\cite{vbench,videoscore}. EWMBench evaluates scene, motion, and semantic quality~\cite{ewmbench}, while WorldArena connects perception to data generation, policy evaluation, and planning~\cite{worldarena}. Framewise appearance alone does not establish action-dependent physical accuracy. We separately evaluate object motion, gripper motion, and video quality, alongside paired-action response differences.

\begin{figure}[t]
\centering
\includegraphics[width=\linewidth]{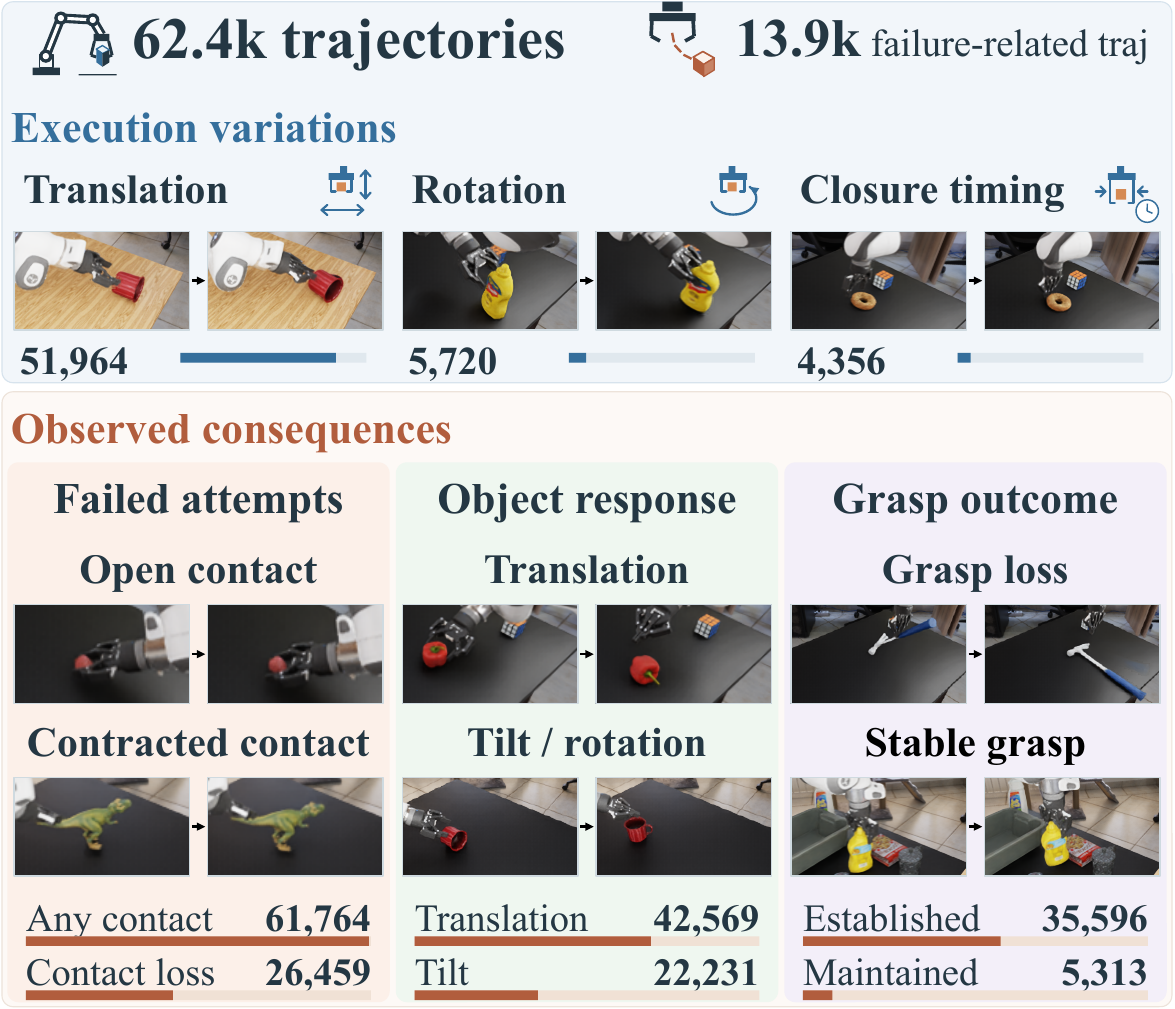}
\caption{Overview of the Reforge dataset. Execution variations span translation, rotation, and closure timing; observed consequences include failed contact, object responses, and grasp outcomes. Consecutive RGB frames illustrate representative changes. Top counts cover the full dataset; bars summarize simulated trajectories, with overlapping output labels.}
\label{fig:dataset}
\end{figure}

\section{Method}
\label{sec:method}

\subsection{Action-Conditioned World Model}
An action-conditioned world model (AC-WM) predicts future observations from visual context and a robot action sequence~\cite{visualforesight,sv2p,unisim}. For physical interaction, the prediction must resolve how contact location, object pose, and gripper state affect the environment's response. We study future RGB prediction conditioned on an initial image and the recorded end-effector poses and gripper states.

Given an initial observation $I_0$ and robot states $s_t^r$, the action adapter $h_b$ of backbone $b$ maps adjacent states to action conditions $a_t^{(b)}$. The resulting conditional distribution over $H_b$ future frames is
\begin{equation}
 p_\theta^{(b)}(I_{1:H_b}\mid I_0,a_{0:H_b-1}^{(b)}),\quad
 a_t^{(b)}=h_b(s_t^r,s_{t+1}^r).
 \label{eq:task}
\end{equation}
The adapter preserves each backbone's action representation within a common prediction task: modeling the visual response to the recorded robot execution.

\begin{figure*}[t]
\centering
\includegraphics[width=\textwidth]{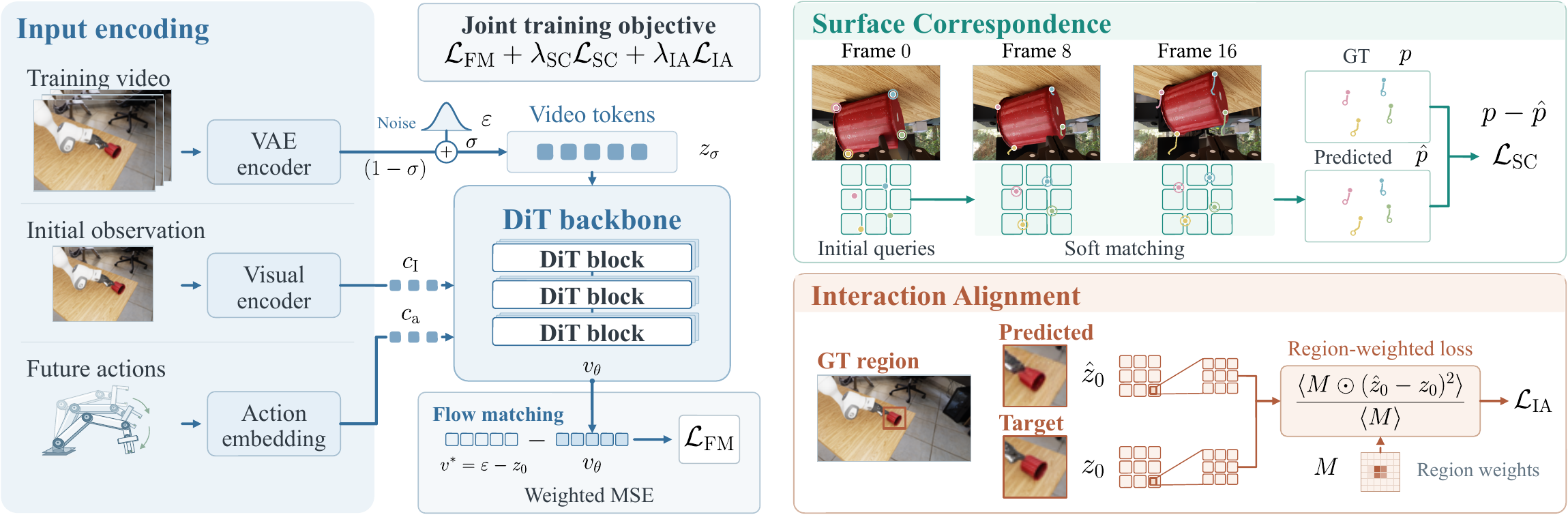}
\caption{WorldReforge augments an action-conditioned DiT with spatial and temporal supervision. The initial observation and future actions condition flow-matching prediction. Interaction Alignment weights reconstruction in interaction regions; Surface Correspondence uses physical GT to align the same surface points across frames through backbone features. RGB examples illustrate regions and tracks. Both auxiliary objectives share the backbone without adding inference-time parameters.}
\label{fig:framework}
\end{figure*}

\subsection{The Reforge dataset}
\label{sec:data}
\subsubsection{Scale and Hierarchical Structure}
\label{sec:data_distribution}
The Reforge dataset integrates diverse interaction experience with synchronized physical ground truth. In simulation, we construct successful reference executions using RoboLab~\cite{robolab} and cuRobo~\cite{curobo}, then perturb end-effector translation, rotation, and gripper closure timing. The resulting successful and failed interactions are recorded with synchronized RGB video, robot motion, and physical states, linking execution conditions to observed responses. \Cref{fig:dataset} summarizes this hierarchy.

The dataset comprises 62,425 trajectories across 106 object assets. Its simulated component contains 62,125 trajectories from 96 assets, including 13,577 failure-related trajectories. Reforge-Real contributes 300 real-world failure trajectories across 10 objects, while Reforge-Eval denotes the simulated evaluation component.

\para{Hierarchical inputs}
Inputs distinguish physical context---object attributes, initial poses, and scene configurations---from execution variations. In the simulated component, execution labels describe end-effector translation (51,964 trajectories), rotation (5,720), and gripper closure timing (4,356), relating robot actions to the conditions under which objects respond.

\para{Hierarchical outputs}
Outputs separate time-aligned contact and object-motion events from grasp outcomes. Event labels record contact formation and loss, object translation, rotation, tilt, and post-grasp slip. Outcome labels indicate whether a grasp is established and maintained. Coexisting events are retained to distinguish transient object responses from sustained holding.

\subsubsection{Organizing Interaction Experience}
\label{sec:data_organization}
\label{sec:selection}
We organize interaction segments by execution variation and response type. Successful grasping and stable holding provide examples of sustained control, whereas failed attempts reveal contact-induced object motion and are grouped by measured gripper state (contracted or open). Because the same execution variation can yield different outcomes, we record execution and response labels separately, retaining the associated object motion and contact events in each segment. Subsequent experiments investigate the complementary value of combining successful reference experience with diverse failed contact.

\subsection{WorldReforge}
\label{sec:worldreforge}
\label{sec:posttraining}
The Reforge dataset supplies complementary interactions together with precise physical ground truth. WorldReforge converts these records into two learning targets: local reconstruction through IA and temporal surface correspondence through SC. As shown in \cref{fig:framework}, both targets supervise the generative backbone as it learns the conditional video distribution in \cref{eq:task}.

Our formulation builds on denoising objectives for image~\cite{ddpm} and video generation~\cite{videodiffusion}, using latent representations~\cite{latentdiffusion} and Transformer backbones~\cite{dit}. Let $z_0$ denote the latent representation of the observed future video and $c=(I_0,a_{0:H-1})$ the observation and action conditions. Flow matching~\cite{flowmatching} constructs a path between data and Gaussian noise $\epsilon$ and learns its conditional velocity field:
\begin{equation}
\begin{aligned}
 z_\sigma&=(1-\sigma)z_0+\sigma\epsilon,\quad v^*=\epsilon-z_0,\\
 \mathcal L_{\rm FM}&=\omega(\sigma)\operatorname{mean}
 \big[(v_\theta(z_\sigma,\sigma,c)-v^*)^2\big].
\end{aligned}
\label{eq:fm_objective}
\end{equation}
Here $\sigma$ is the noise level, $\omega(\sigma)$ is the backbone-specific training weight, and the mean covers valid prediction positions.

\subsubsection{Interaction Alignment}
\label{sec:alignment}
IA concentrates the reconstruction signal on object appearance and position in interaction regions. Object states, geometry, and camera projections define a soft mask $M$ covering objects and their contact neighborhoods. From the clean-future estimate $\hat z_0=z_\sigma-\sigma v_\theta$, IA aligns predicted and target video latents within these regions:
\begin{equation}
 \mathcal L_{\rm IA}=
 \frac{\sum_{t,p}M_{t,p}\operatorname{mean}_c[(\hat z_{0,t,p,c}-z_{0,t,p,c})^2]}
 {\sum_{t,p}M_{t,p}}.
 \label{eq:roi}
\end{equation}
Normalization by total mask weight gives the interaction region a dedicated learning signal even when it occupies a small fraction of the image.

The linear flow path makes IA a spatially focused extension of the generative objective:
\begin{equation}
\begin{aligned}
 \hat z_0-z_0&=\sigma(v^*-v_\theta),\\
 \mathcal L_{\rm IA}&=\sigma^2\big\langle(v_\theta-v^*)^2\big\rangle_M.
\end{aligned}
\label{eq:ia_flow}
\end{equation}
The operator $\langle\cdot\rangle_M$ denotes the region-and-channel average in \cref{eq:roi}.

\subsubsection{Surface Correspondence}
\label{sec:correspondence}
SC supervises motion through cross-frame correspondences of physical surface points. Each point $X_i$ is initially visible and fixed in the object coordinate system, preserving its physical identity as the object moves. The measured object transform $T_t^o$, camera transform $T_t^c$, and intrinsics $K_t$ determine its projected trajectory:
\begin{equation}
 p_{it}^{\rm GT}=\pi\!\left(K_t(T_t^c)^{-1}T_t^oX_i\right).
 \label{eq:gt_projection}
\end{equation}
Scene and robot geometry determine visibility. SC operates directly on intermediate backbone features. For each initial surface point, SC interpolates a query feature and matches it to future wrist-view features with a softmax over cosine similarities. The predicted position $\hat p_{it}$ is the probability-weighted mean of candidate locations. Let $m_{it}$ mask points that are visible and inside the valid feature domain, and let $D_w$ normalize wrist-view dimensions:
\begin{equation}
 \mathcal L_{\rm SC}=
 \sqrt{\frac{\sum_{i,t}m_{it}\|D_w^{-1}(\hat p_{it}-p_{it}^{\rm GT})\|_2^2}
 {2\sum_{i,t}m_{it}}+10^{-6}}.
 \label{eq:tracking}
\end{equation}
The projected trajectories account for object and camera motion. They train the generative features to locate the same physical surface across frames, providing temporal supervision that complements IA's spatial reconstruction target.

The full WorldReforge objective combines flow matching with IA and SC over the sampling distribution $q_{\mathcal S}$ defined in \cref{sec:data_organization}. With an auxiliary-gradient ramp $r(s)$ and valid noise-range and visibility masks $g$, the objective is
\begin{equation}
 \mathcal L_\theta=\mathbb E\big[\mathcal L_{\rm FM}+r(s)(\lambda_R\mathcal L_{\rm IA}^{g}+\lambda_T\mathcal L_{\rm SC}^{g})\big].
\label{eq:loss}
\end{equation}
The expectation averages over segments $w\sim q_{\mathcal S}$, noise levels $\sigma$, and Gaussian noise $\epsilon$.
IA and SC train the shared backbone without adding inference parameters or changing its architecture. Reforge denotes post-training on the Reforge dataset with the base flow-matching objective; WorldReforge adds IA and SC under matched data and training budgets.

\begin{table*}[t]
\centering\fontsize{8}{9.2}\selectfont\compacttable
\caption{Quantitative results for five pretrained backbones from three model families, with post-training comparisons on Cosmos3 Nano, DreamDojo, and LingBot-VA. Pretrained denotes the initial checkpoint; Reforge uses the base objective on the Reforge dataset, while WorldReforge adds IA and SC. Object and gripper motion are evaluated separately on Reforge-Eval; video quality is reported on Reforge-Eval, DROID, and Reforge-Real. Among learned models, bold marks column-best Cosmos3 Nano WorldReforge values; other best values and runners-up are underlined.}
\label{tab:main}
\setlength{\tabcolsep}{1.25pt}
\begin{tabular*}{\linewidth}{@{\extracolsep{\fill}}llrrrrrrrrrrrrr@{}}\toprule
 & & \multicolumn{7}{c}{Reforge-Eval} & \multicolumn{3}{c}{DROID} & \multicolumn{3}{c}{Reforge-Real}\\
\cmidrule(lr){3-9}\cmidrule(lr){10-12}\cmidrule(lr){13-15}
 & & \multicolumn{2}{c}{Object motion} & \multicolumn{2}{c}{Gripper motion} & \multicolumn{3}{c}{Video quality} & \multicolumn{3}{c}{Video quality} & \multicolumn{3}{c}{Video quality}\\
\cmidrule(lr){3-4}\cmidrule(lr){5-6}\cmidrule(lr){7-9}\cmidrule(lr){10-12}\cmidrule(lr){13-15}
Backbone & Method & ADE$\downarrow$ & FDE$\downarrow$ & ADE$\downarrow$ & FDE$\downarrow$ & PSNR$\uparrow$ & SSIM$\uparrow$ & $Q\uparrow$ & PSNR$\uparrow$ & SSIM$\uparrow$ & $Q\uparrow$ & PSNR$\uparrow$ & SSIM$\uparrow$ & $Q\uparrow$\\\midrule
Static & \NA & 9.38 & 14.42 & 9.47 & 14.22 & 21.45 & 0.761 & \NA & 17.61 & 0.719 & \NA & 19.86 & 0.861 & \NA \\
\midrule
DreamDojo~\cite{dreamdojo} & Pretrained & 9.57 & 14.64 & 9.54 & 14.27 & 18.25 & 0.685 & 63.08 & 17.27 & 0.691 & 88.00 & 19.41 & 0.824 & 90.00 \\
LingBot-VA~\cite{lingbot} & Pretrained & 9.60 & 14.69 & 9.63 & 14.38 & 20.77 & 0.691 & 91.92 & 17.09 & 0.674 & 68.00 & 19.19 & 0.810 & 83.00 \\
Cosmos3 Edge~\cite{cosmos3} & Pretrained & 8.40 & 13.30 & 7.42 & 9.86 & 21.44 & 0.743 & \underline{99.67} & 19.49 & 0.747 & \underline{100.00} & 20.47 & 0.847 & \underline{100.00} \\
Cosmos3 Nano~\cite{cosmos3} & Pretrained & 9.07 & 14.65 & 5.87 & 6.59 & 21.99 & 0.758 & 94.50 & 20.38 & 0.777 & \underline{100.00} & 20.80 & 0.860 & 99.00 \\
Cosmos3 Super~\cite{cosmos3} & Pretrained & 8.63 & 13.94 & 6.32 & 7.67 & 22.24 & 0.766 & 96.25 & 20.82 & 0.786 & 98.50 & 20.91 & 0.862 & \underline{100.00} \\
\midrule
DreamDojo & Reforge & 11.81 & 17.02 & 7.62 & 11.12 & 20.36 & 0.716 & 97.42 & 17.03 & 0.694 & 98.00 & 19.41 & 0.846 & \underline{100.00} \\
LingBot-VA & Reforge & 7.12 & 12.06 & 2.28 & 4.85 & 24.14 & 0.782 & 91.08 & 18.25 & 0.681 & 74.00 & 21.11 & 0.821 & 58.00 \\
Cosmos3 Nano & Reforge & \underline{6.81} & \underline{10.57} & \underline{2.18} & \underline{3.03} & \underline{25.68} & \underline{0.854} & 99.58 & \underline{21.32} & \underline{0.796} & 99.50 & \underline{21.31} & \underline{0.866} & 97.00 \\
\midrule
DreamDojo & WorldReforge & 11.62 & 16.65 & 7.61 & 10.96 & 20.40 & 0.718 & 97.42 & 17.03 & 0.694 & 98.00 & 19.42 & 0.846 & \underline{100.00} \\
LingBot-VA & WorldReforge & 7.12 & 11.94 & 2.25 & 4.85 & 24.17 & 0.782 & 93.42 & 18.40 & 0.685 & 74.00 & 21.25 & 0.825 & 58.00 \\
Cosmos3 Nano & WorldReforge & \textbf{6.51} & \textbf{10.05} & \textbf{2.12} & \textbf{2.90} & \textbf{25.85} & \textbf{0.858} & \textbf{99.67} & \textbf{21.42} & \textbf{0.797} & \textbf{100.00} & \textbf{21.48} & \textbf{0.867} & \textbf{100.00} \\
\bottomrule\end{tabular*}
\end{table*}

\begin{table*}[t]
\centering
\begin{minipage}[t]{0.485\textwidth}
\centering\fontsize{7.5}{8.7}\selectfont\compacttable
\caption{Quantitative Results on seen and unseen object assets in Reforge-Eval with Cosmos3 Nano. Seen and unseen are defined relative to the post-training data.}
\label{tab:unseen_assets}
\setlength{\tabcolsep}{0.6pt}
\begin{tabular*}{\linewidth}{@{\extracolsep{\fill}}lrrrrrr@{}}\toprule
 & \multicolumn{3}{c}{Seen assets} & \multicolumn{3}{c}{Unseen assets}\\
\cmidrule(lr){2-4}\cmidrule(lr){5-7}
Method & O.ADE$\downarrow$ & G.ADE$\downarrow$ & PSNR$\uparrow$ & O.ADE$\downarrow$ & G.ADE$\downarrow$ & PSNR$\uparrow$\\\midrule
Pretrained & 8.65 & 5.70 & 21.82 & 9.86 & 6.19 & 22.32 \\
Reforge & \underline{6.89} & \underline{2.17} & \underline{25.18} & \underline{6.67} & \underline{2.20} & \underline{26.61} \\
WorldReforge & \textbf{6.57} & \textbf{2.12} & \textbf{25.37} & \textbf{6.41} & \textbf{2.10} & \textbf{26.76} \\
\bottomrule\end{tabular*}
\end{minipage}\hfill
\begin{minipage}[t]{0.485\textwidth}
\centering\fontsize{7.5}{8.7}\selectfont\compacttable
\caption{Quantitative Results on failure-related and non-failure-related clips in Reforge-Eval with Cosmos3 Nano. The non-failure-related group includes successful holding.}
\label{tab:failure_groups}
\setlength{\tabcolsep}{0.6pt}
\begin{tabular*}{\linewidth}{@{\extracolsep{\fill}}lrrrrrr@{}}\toprule
 & \multicolumn{3}{c}{Failure-related} & \multicolumn{3}{c}{Non-failure-related}\\
\cmidrule(lr){2-4}\cmidrule(lr){5-7}
Method & O.ADE$\downarrow$ & G.ADE$\downarrow$ & PSNR$\uparrow$ & O.ADE$\downarrow$ & G.ADE$\downarrow$ & PSNR$\uparrow$\\\midrule
Pretrained & 9.44 & 5.48 & 21.92 & 8.67 & 6.29 & 22.07 \\
Reforge & \underline{7.82} & \underline{2.15} & \underline{25.39} & \underline{5.73} & \underline{2.21} & \underline{25.98} \\
WorldReforge & \textbf{7.48} & \textbf{2.08} & \textbf{25.56} & \textbf{5.47} & \textbf{2.15} & \textbf{26.16} \\
\bottomrule\end{tabular*}
\end{minipage}
\end{table*}

\newcommand{\delayedablationtables}{%
\begin{table*}[t]
\centering\fontsize{8}{9.2}\selectfont\compacttable
\caption{Effect of execution-experience composition on Cosmos3 Nano. Rot. + Clos., Trans. + Clos., and Trans. + Rot. each allocate 50\% of the training data to each named execution type. Trans. + Rot. + Clos. assigns one third to translation, rotation, and closure timing.}
\label{tab:input_ablation}
\setlength{\tabcolsep}{0.5pt}
\begin{tabular*}{\linewidth}{@{\extracolsep{\fill}}lcccccccccc@{}}\toprule
 & \multicolumn{2}{c}{Translation} & \multicolumn{2}{c}{Rotation} & \multicolumn{2}{c}{Closure} & \multicolumn{2}{c}{Reforge-Eval} & DROID & Reforge-Real\\
\cmidrule(lr){2-3}\cmidrule(lr){4-5}\cmidrule(lr){6-7}\cmidrule(lr){8-9}\cmidrule(lr){10-11}
Experience & O.ADE/G.ADE$\downarrow$ & PSNR$\uparrow$ & O.ADE/G.ADE$\downarrow$ & PSNR$\uparrow$ & O.ADE/G.ADE$\downarrow$ & PSNR$\uparrow$ & O.ADE/G.ADE$\downarrow$ & PSNR$\uparrow$ & PSNR$\uparrow$ & PSNR$\uparrow$\\\midrule
Rot. + Clos. & \underline{5.39}/2.88 & \underline{25.08} & \underline{6.91}/3.01 & \underline{24.25} & \textbf{5.55}/3.36 & \underline{25.24} & 7.35/3.02 & 24.42 & 21.19 & 21.26 \\
Trans. + Clos. & 5.40/\underline{2.61} & 25.05 & 7.03/2.81 & 24.16 & 5.68/\underline{3.04} & 25.14 & \underline{7.29}/\underline{2.77} & 24.38 & 21.18 & 21.32 \\
Trans. + Rot. & \textbf{5.30}/2.78 & \textbf{25.15} & \textbf{6.85}/\underline{2.78} & \textbf{24.33} & \underline{5.61}/3.25 & \textbf{25.26} & \textbf{7.25}/2.84 & \textbf{24.53} & \textbf{21.24} & \textbf{21.37} \\
Trans. + Rot. + Clos. & 5.58/\textbf{2.54} & 25.07 & 7.03/\textbf{2.66} & \underline{24.25} & 5.78/\textbf{3.00} & 25.21 & 7.38/\textbf{2.68} & \underline{24.45} & \underline{21.22} & \underline{21.36} \\
\bottomrule\end{tabular*}
\end{table*}

\begin{table*}[t]
\centering\fontsize{8}{9.2}\selectfont\compacttable
\caption{Complementarity of successful holding and failed contact on Cosmos3 Nano. S uses 100\% successful-holding data. S + Contracted and S + Open each combine 50\% successful holding with 50\% contracted-gripper or open-gripper contact, respectively. S + Contracted + Open uses 50\% successful holding and 25\% of each failed-contact type.}
\label{tab:outcome_ablation}
\setlength{\tabcolsep}{0.5pt}
\begin{tabular*}{\linewidth}{@{\extracolsep{\fill}}lcccccccccc@{}}\toprule
 & \multicolumn{2}{c}{Contracted-contact} & \multicolumn{2}{c}{Open-contact} & \multicolumn{2}{c}{Successful holding} & \multicolumn{2}{c}{Reforge-Eval} & DROID & Reforge-Real\\
\cmidrule(lr){2-3}\cmidrule(lr){4-5}\cmidrule(lr){6-7}\cmidrule(lr){8-9}\cmidrule(lr){10-11}
Experience & O.ADE/G.ADE$\downarrow$ & PSNR$\uparrow$ & O.ADE/G.ADE$\downarrow$ & PSNR$\uparrow$ & O.ADE/G.ADE$\downarrow$ & PSNR$\uparrow$ & O.ADE/G.ADE$\downarrow$ & PSNR$\uparrow$ & PSNR$\uparrow$ & PSNR$\uparrow$\\\midrule
S & 12.78/3.08 & 23.31 & 7.21/3.07 & 24.29 & \textbf{7.09}/\textbf{2.71} & \textbf{24.59} & 7.70/3.33 & 24.30 & 21.05 & 21.23 \\
S + Contracted & \textbf{10.48}/\textbf{2.29} & \textbf{24.06} & \textbf{6.27}/\underline{2.43} & 24.72 & 7.33/\underline{2.77} & \underline{24.38} & \textbf{7.06}/\underline{2.72} & \underline{24.62} & \underline{21.19} & \textbf{21.37} \\
S + Open & 11.82/2.71 & 23.41 & 6.71/2.54 & \textbf{25.12} & \underline{7.16}/2.89 & 24.33 & 7.31/3.10 & 24.52 & 21.02 & 21.20 \\
S + Contracted + Open & \underline{11.22}/\underline{2.38} & \underline{23.78} & \underline{6.49}/\textbf{2.16} & \underline{25.08} & 7.59/2.85 & 24.27 & \underline{7.25}/\textbf{2.64} & \textbf{24.67} & \textbf{21.24} & \underline{21.24} \\
\bottomrule\end{tabular*}
\end{table*}

}
\afterpage{\qualitativefigure}
\afterpage{\afterpage{\afterpage{\delayedablationtables}}}

\newcommand{\qualitativefigure}{%
\begin{figure*}[t]
\centering
\includegraphics[width=\textwidth]{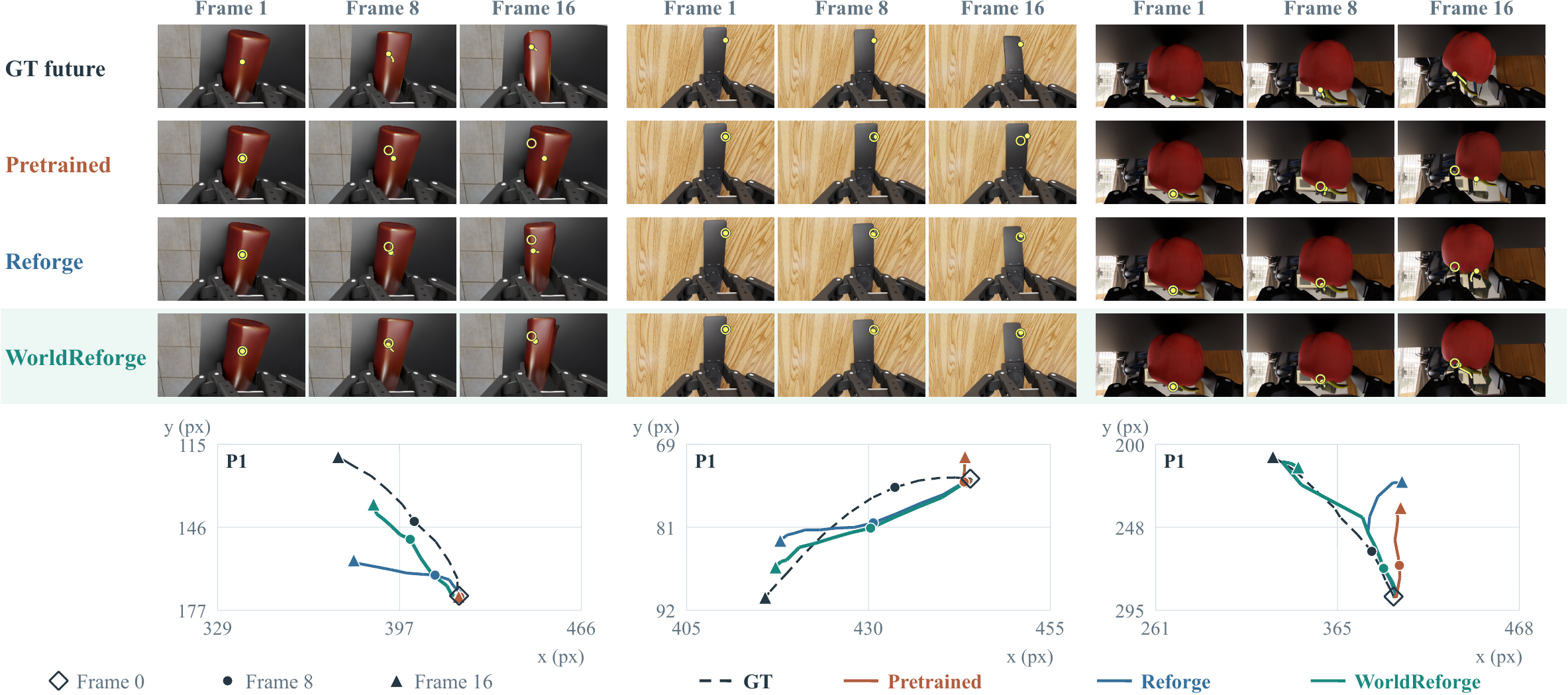}
\caption{Interaction responses and object motion with Cosmos3 Nano. Top: RGB sequences compare recorded futures with Pretrained, Reforge, and WorldReforge predictions at Frames 1, 8, and 16. Yellow markers identify the common surface point P1; rings indicate its GT position. Bottom: image-plane $x$--$y$ trajectories of P1 across all frames, with dashed simulator GT and solid CoTracker tracks from generated videos. Each plot uses equal $x$--$y$ pixel scales; diamonds, circles, and triangles denote Frames 0, 8, and 16.}
\label{fig:qualitative}
\end{figure*}
}

\section{Experiments}
\label{sec:experiments}
\subsection{Experimental Setup}
\para{Datasets}
We post-train on 25.6k interaction segments from the Reforge dataset, nearly 50\% drawn from failed trajectories. Evaluation covers 1,500 interaction clips from Reforge-Eval, DROID~\cite{droid}, and Reforge-Real. Training and Reforge-Eval share some object assets but no trajectories or initial states. Real-world videos are used only for evaluation.

\para{Baselines}
We evaluate five backbones from three model families. Cosmos3~\cite{cosmos3,cosmosmatrix} unifies visual, language, and action modeling; we use Cosmos3 Edge, Cosmos3 Nano, and Cosmos3 Super. DreamDojo~\cite{dreamdojo} transfers interaction knowledge from large-scale egocentric human video to robots through continuous latent actions. LingBot-VA~\cite{lingbot} uses autoregressive diffusion to jointly model future frames and robot control with shared video and action representations.

\para{Metrics}
We use a $640\times544$ canvas combining wrist, left-shoulder, and right-shoulder views. CoTracker~\cite{cotracker3} tracks first-frame points in each generated view; we compare these tracks with projected simulator ground truth over frames 1--16.
ADE and FDE measure average and final-frame position errors, respectively. We score objects and grippers separately, using the mean Euclidean error over GT-visible points for each entity, denoted $\bar e_{n,t}^{(s)}$:
\begin{equation}
\begin{aligned}
\operatorname{ADE}^{(s)}&=\frac{1}{|\mathcal N_A^{(s)}|}\sum_{n\in\mathcal N_A^{(s)}}
 \frac{1}{|\mathcal T_n^{(s)}|}\sum_{t\in\mathcal T_n^{(s)}}\bar e_{n,t}^{(s)},\\
\operatorname{FDE}^{(s)}&=\frac{1}{|\mathcal N_F^{(s)}|}\sum_{n\in\mathcal N_F^{(s)}}\bar e_{n,16}^{(s)}.
\end{aligned}
\label{eq:motion_metrics}
\end{equation}
Here $s$ denotes the object or gripper, and $\mathcal T_n^{(s)}$ indexes valid future frames. The sets $\mathcal N_A^{(s)}$ and $\mathcal N_F^{(s)}$ contain videos with valid ADE and FDE references, respectively. PSNR and SSIM~\cite{ssim} measure reconstruction quality over the full canvas. Motion and reconstruction metrics weight videos equally. The Visual Quality Pass Rate, $Q$, is the proportion of videos without evident visual abnormalities, assessed independently by Qwen3.8-27B. Throughout, ADE/FDE are reported in pixels, PSNR in dB, and $Q$ as a percentage; SSIM is unitless. In tables, O. and G. denote object and gripper.

\subsection{Quantitative Results}
\label{sec:scale}
\para{More accurate physical responses}
\Cref{tab:main} separates the value of interaction experience from the additional contribution of physical supervision through Pretrained, Reforge, and WorldReforge. Cosmos3 Nano WorldReforge achieves the lowest errors on all four motion metrics. It reduces object ADE by 28.2\% relative to Pretrained and by 4.4\% relative to Reforge under matched data and budget, improving both object trajectories and final positions. Overall PSNR increases by 3.27\,dB over Pretrained.

\para{Learning from failure-enriched experience}
Base post-training on the Reforge dataset reduces object ADE by 24.9\% on Cosmos3 Nano and 25.8\% on LingBot-VA relative to Pretrained. Both also improve gripper prediction and Reforge-Eval reconstruction. DreamDojo gains 2.12\,dB in Reforge-Eval PSNR and lowers gripper ADE from 9.54 to 7.62 pixels. These gains highlight the value of failure-enriched interaction experience even under the base generative objective.

\para{Physical supervision improves motion prediction}
WorldReforge improves object and gripper motion prediction over Reforge across all three post-trained model families. Cosmos3 Nano's object FDE decreases by 5.0\%, and DreamDojo's object ADE/FDE by 1.6\%/2.2\%. The dataset supplies interaction responses; IA and SC improve their prediction through additional spatial and temporal learning signals.

\para{Generalization to real-world video}
\label{sec:real}
Cosmos3 Nano WorldReforge increases PSNR over Pretrained by 1.04\,dB on DROID and 0.68\,dB on Reforge-Real, with higher SSIM in both domains. Relative to Reforge, LingBot-VA improves PSNR and SSIM in both real domains, and DreamDojo also improves Reforge-Real PSNR. Post-training uses simulated interactions alone, yet improves reconstruction in both real-world domains, supporting sim-to-real generalization in visual prediction.

\para{Visual quality}
Cosmos3 Nano WorldReforge attains $Q$ of 99.67\%, 100.00\%, and 100.00\% on Reforge-Eval, DROID, and Reforge-Real; DreamDojo's Reforge-Eval pass rate increases from 63.08\% to 97.42\%.

\para{Seen and unseen object assets}
\Cref{tab:unseen_assets} evaluates approximately 800 clips of assets seen in post-training and 400 of unseen assets. WorldReforge outperforms Reforge in both groups; on unseen assets, object ADE falls by 35.02\% and PSNR rises by 4.44\,dB relative to Pretrained.

\para{Failure-related and other interactions}
Failure-related clips account for approximately half of the evaluation set in \cref{tab:failure_groups}. Relative to Reforge, WorldReforge lowers object ADE by 4.36\% on failure-related and 4.52\% on non-failure-related clips, with higher PSNR in both groups.

\subsection{Qualitative Results}
\label{sec:qualitative}
\Cref{fig:qualitative} shows Cosmos3 Nano predictions of a rotating sauce bottle, a tilting remote control, and a moving red pepper. Pretrained predictions remain near the initial pose, Reforge recovers the main motion trend, and WorldReforge follows the recorded direction, displacement, and final pose more closely.

\section{Analysis}
\label{sec:analysis}
We analyze how execution coverage, complementary successful and failed experience, action sensitivity, and physical supervision shape interaction prediction.

\subsection{Execution Coverage}
\label{sec:input_ablation}
Translation, rotation, and closure timing expose different aspects of object--gripper interaction. By varying their proportions in \cref{tab:input_ablation}, we examine how these experiences contribute to object and gripper prediction, both within and across execution types. Each execution type is evaluated on the corresponding 200-video subset of Reforge-Eval.

\para{Object-response prediction}
Combining translation and rotation gives the lowest object ADE under translation and rotation changes and across Reforge-Eval (7.25 pixels). It also achieves the highest PSNR on Reforge-Eval and both real domains. The rotation and closure recipe instead gives the lowest object ADE under closure changes, demonstrating the distinct value of different execution experiences.

\para{Complementary execution experience}
Including closure alongside translation and rotation yields the lowest gripper ADE across all execution conditions, reaching 2.68 pixels overall. Together with the object-response results, this supports emphasizing translation and rotation while retaining closure experience for accurate gripper dynamics. These findings highlight the different contributions of execution types to object and gripper motion prediction.

\subsection{Failure Complementarity}
\label{sec:outcome_ablation}
We combine successful holding with contracted-gripper contact, open-gripper contact, or both to examine their complementarity in \cref{tab:outcome_ablation}. The two contact groups are distinguished by measured gripper state during failed attempts. Each response group is evaluated on the corresponding 200-video subset of Reforge-Eval.

\para{Failed contact improves object-response prediction}
Adding contracted-gripper contact reduces object ADE in that group from 12.78 to 10.48 pixels, an 18.0\% reduction, and yields the best gripper ADE and PSNR in that condition. It also reduces object ADE by 13.0\% under open-gripper contact and gives the lowest object ADE across Reforge-Eval. Adding open-gripper contact raises PSNR in the corresponding group by 0.83\,dB while reducing both motion errors.

\para{Diverse failed contact benefits both failure groups}
At the same 50\% failure-experience share, mixing contracted and open contact reduces object ADE by 12.2\% and 9.9\% in the respective conditions relative to success-only training. The mixture also gives the lowest overall gripper ADE (2.64 pixels) and highest Reforge-Eval and DROID PSNR. These comparisons suggest that contact-state diversity matters even when the overall failure-data proportion is fixed.

\para{Successful holding anchors state persistence}
Success-only training gives the best object ADE, gripper ADE, and PSNR under holding. Successful and failed experience therefore serve complementary purposes: holding supplies examples of a maintained grasp, while failed contact supplies examples of contact-induced change.

\subsection{Action-Dependent Responses}
\label{sec:paired_response}

\begin{table}[!t]
\centering\fontsize{8.2}{9.4}\selectfont\compacttable
\caption{Action-dependent response prediction by Cosmos3 Nano on 100 execution pairs. Each pair shares an initial observation but uses different actions. RDE measures error in the predicted difference between the two futures and is displayed $\times10^3$.}
\label{tab:paired_response}
\setlength{\tabcolsep}{1.0pt}
\begin{tabular*}{\linewidth}{@{\extracolsep{\fill}}lrrrr@{}}\toprule
Method & RDE$\downarrow$ & O.ADE$\downarrow$ & G.ADE$\downarrow$ & PSNR$\uparrow$\\\midrule
Static & 6.0556 & 9.16 & 8.52 & 22.31 \\
\midrule
Pretrained & 7.3521 & 10.68 & 5.87 & 22.18 \\
Reforge & \underline{4.3418} & \underline{8.30} & \underline{2.37} & \underline{25.97} \\
WorldReforge & \textbf{4.1717} & \textbf{8.02} & \textbf{2.31} & \textbf{26.06} \\
\bottomrule\end{tabular*}
\end{table}

\begin{figure}[!t]
\centering
\includegraphics[width=\linewidth]{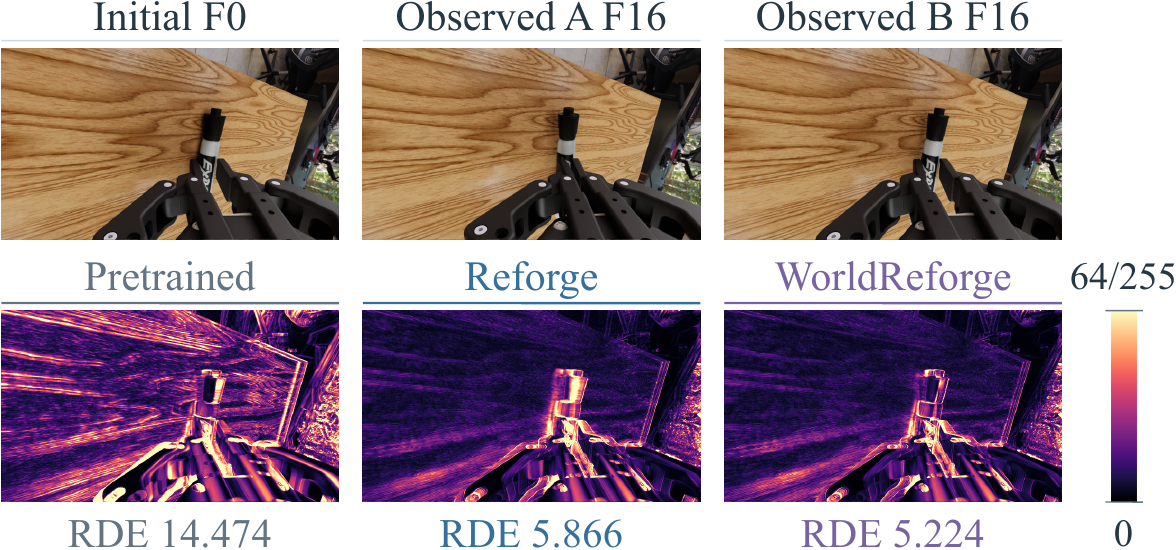}
\caption{Different actions produce different physical responses: the marker slips in execution A and is lifted in B. Top: shared Frame 0 and recorded Frame 16 outcomes. Bottom: wrist-view RMS maps of the RGB branch-difference residual over all future frames and channels; darker indicates lower error. RDE for this pair's wrist view is displayed $\times10^3$.}
\label{fig:paired_response}
\end{figure}
Faithful action-conditioned prediction requires capturing how physical responses change with the robot's execution. We compare pairs that share an initial observation but follow different action sequences, measuring how closely the predicted response difference matches the recorded one. For observed branch futures $I^A,I^B$ and their predictions $\hat I^A,\hat I^B$, we define Response Difference Error (RDE) as the mean squared residual between predicted and observed RGB branch differences:
\begin{equation}
 E_{\Delta}=\operatorname{mean}\!\left[
 \big((\hat I^B-\hat I^A)-(I^B-I^A)\big)^2\right].
 \label{eq:paired_response}
\end{equation}
We first average the squared RGB residual across all pixels and future frames of each pair, then average across pairs. Object/gripper ADE and PSNR evaluate each predicted future against its corresponding recording.

\Cref{tab:paired_response} shows that WorldReforge reduces response difference error by 43.3\% relative to Pretrained and by 3.9\% relative to Reforge on these pairs. Relative to Reforge, object ADE falls by 3.4\% and gripper ADE by 2.9\%, while PSNR increases by 0.09\,dB. WorldReforge improves each branch's motion prediction and the difference between the two generated responses, linking trajectory accuracy to sensitivity to the action input.

In \cref{fig:paired_response}, the marker slips during the grasp attempt in execution A, while execution B successfully lifts it. WorldReforge reduces the residual around the object and contact region, more accurately reproducing the observed change between futures.

\subsection{Ablation on Physical Supervision}
\label{sec:supervision}
\begin{table}[!ht]
\centering\fontsize{7.5}{8.7}\selectfont\compacttable
\caption{Ablation on physical supervision for Cosmos3 Nano. IA emphasizes interaction-region reconstruction, while SC supervises cross-frame surface motion.}
\label{tab:supervision}
\setlength{\tabcolsep}{1.3pt}
\begin{tabular*}{\linewidth}{@{\extracolsep{\fill}}lrrrrr@{}}\toprule
 & \multicolumn{2}{c}{Reforge-Eval} & \multicolumn{3}{c}{PSNR$\uparrow$}\\
\cmidrule(lr){2-3}\cmidrule(lr){4-6}
Objective & O.ADE$\downarrow$ & G.ADE$\downarrow$ & Reforge-Eval & DROID & Reforge-Real\\\midrule
Reforge & 7.02 & 2.85 & 24.53 & 21.07 & 21.28 \\
+ IA & 6.84 & \underline{2.84} & \underline{24.58} & 21.08 & \textbf{21.30} \\
+ SC & \underline{6.81} & \underline{2.84} & 24.55 & \underline{21.15} & 21.28 \\
WorldReforge & \textbf{6.59} & \textbf{2.66} & \textbf{24.67} & \textbf{21.18} & \textbf{21.30} \\
\bottomrule\end{tabular*}
\end{table}
\Cref{tab:supervision} isolates the individual and joint benefits of IA and SC. All four variants are post-trained on a 6.4k-segment subset of the Reforge dataset. Joint supervision lowers object ADE by 3.2\% and gripper ADE by 6.3\% relative to the stronger individual objective for each metric. It achieves the highest PSNR on Reforge-Eval and DROID and ties with IA on Reforge-Real. These gains highlight the complementarity of interaction-region reconstruction and cross-frame surface correspondence. Both objectives share the generative backbone without introducing additional inference-time parameters.

\section{Conclusion}
\label{sec:discussion}
WorldReforge strengthens action-conditioned world models through complementary interaction experience and physical supervision. The Reforge dataset connects execution variations to physical outcomes, while Interaction Alignment and Surface Correspondence focus learning on contact regions and cross-frame surface motion. This connects global video reconstruction with the dynamics of physical interaction. Our results show that failed contact enriches object-response prediction and successful holding supports state persistence. These complementary signals improve interaction prediction, real-world video reconstruction, and sensitivity to action inputs. Future work will extend this framework to longer-horizon manipulation and study how consequence prediction can support closed-loop robot control.

\section*{LLM Usage Statement}
During this work, ChatGPT and Codex assisted with programming, manuscript preparation, and language refinement. The authors checked the resulting code and text against the implementation and experimental evidence, and are responsible for the final paper and accompanying materials.
\balance
\section*{Reproducibility Statement}
We will open-source the data, the trained checkpoints, the models, and the data pipeline upon publication. The action-conditioned model is formulated in \cref{eq:task}, and the physical-supervision objectives are specified in \cref{sec:worldreforge}. The supplementary material provides data-collection details, training and evaluation settings, and additional analyses.

\label{LastMainPage}
\clearpage
\nobalance
\label{ReferencesStart}
{\small\bibliographystyle{ieeenat_fullname}\bibliography{references}}
\clearpage
\twocolumn[{%
  \centering
  {\Large\bfseries Supplementary Material\par}
  \vspace{16pt}%
}]
\pdfbookmark[0]{Supplementary Material}{supplementary}
\raggedbottom
\appendix
\setcounter{table}{0}
\setcounter{figure}{0}
\setcounter{equation}{0}
\renewcommand{\thetable}{S\arabic{table}}
\renewcommand{\thefigure}{S\arabic{figure}}
\renewcommand{\theequation}{S\arabic{equation}}
\renewcommand{\theHtable}{supp.\arabic{table}}
\renewcommand{\theHfigure}{supp.\arabic{figure}}
\renewcommand{\theHequation}{supp.\arabic{equation}}
\renewcommand{\tablename}{Tab.}
\renewcommand{\figurename}{Fig.}
\crefname{table}{Tab.}{Tab.}
\Crefname{table}{Tab.}{Tab.}
\crefname{figure}{Fig.}{Fig.}
\Crefname{figure}{Fig.}{Fig.}
\crefname{section}{Sec.}{Secs.}
\Crefname{section}{Sec.}{Secs.}
\crefname{subsection}{Sec.}{Secs.}
\Crefname{subsection}{Sec.}{Secs.}
\crefname{appendix}{Appendix}{Appendices}
\Crefname{appendix}{Appendix}{Appendices}

\noindent \Cref{app:data} describes dataset construction and real-world capture; \cref{app:setup} gives training settings, evaluation protocols, and the visual-quality prompt; \cref{app:results} presents quantitative analyses; \cref{app:paired} details paired-action evaluation; and \cref{app:qualitative} provides multi-view qualitative results. Supplementary figures, tables, and equations use the prefix S.
\section{The Reforge Dataset}
\label{app:data}
\subsection{Dataset Composition}
The Reforge dataset links variations in robot actions to observed physical consequences. It comprises 62,125 simulated trajectories and 300 real-world trajectories spanning 106 object assets. The simulated data support post-training and evaluation, while the separately collected Reforge-Real data assess prediction in real-world interactions. \Cref{tab:dataset_scale} summarizes both components and their failure-related trajectories.

\begin{table}[ht]
\centering\tablefont
\caption{Dataset composition. All 300 real-world trajectories in Reforge-Real are failure-related.}
\label{tab:dataset_scale}
\begin{tabular*}{\linewidth}{@{\extracolsep{\fill}}lrrr@{}}\toprule
Component & Trajectories & Assets & Failure-related\\\midrule
Simulation & 62,125 & 96 & 13,577\\
Reforge-Real & 300 & 10 & 300\\\midrule
Total & 62,425 & 106 & 13,877\\\bottomrule
\end{tabular*}
\end{table}

\subsection{Reference Executions and Perturbations}
We construct successful reference executions using RoboLab~\cite{robolab} for robot simulation and cuRobo~\cite{curobo} for collision-aware motion planning. Perturbing end-effector translation, rotation, and gripper closure timing produces diverse contact processes and manipulation outcomes. Each execution records synchronized intervention commands, robot motion, and object states, linking action variations to their observed responses.

Each intervention is specified by its reference frame, execution phase, direction, magnitude, and temporal profile. Translation primarily varies the horizontal end-effector position, and rotation primarily varies its orientation about the vertical axis; translations and rotations in other spatial directions are also included.

\begin{table}[ht]
\centering\tablefont
\caption{Execution categories of simulated trajectories. Each trajectory has one primary execution category and may carry multiple physical outcome labels.}
\label{tab:input_counts}
\begin{tabular*}{\linewidth}{@{\extracolsep{\fill}}lr@{}}\toprule
Primary execution category & Trajectories\\\midrule
Translation & 51,964\\
Rotation & 5,720\\
Closure timing & 4,356\\
Unperturbed reference & 85\\\midrule
Total & 62,125\\\bottomrule
\end{tabular*}
\end{table}

Object properties, initial poses, and scene configurations define the physical context of each intervention. These contextual attributes are annotated separately from execution variations, allowing the dataset to describe how the same action variation produces different responses across objects and initial states.

\subsection{Hierarchical Physical Annotations}
Output annotations describe interactions at three levels: contact events, object responses, and grasp outcomes. Contact formation and loss characterize robot--object contact; translation, rotation, tilt, and relative slip describe object motion; grasp outcomes indicate whether a grasp is established and maintained during observation. These annotations are temporally aligned and may coexist: an object can rotate while remaining grasped, or move under contact without being lifted.

\begin{table}[!htbp]
\centering\tablefont
\caption{Physical annotations of simulated trajectories. Contact, motion, and grasp labels may overlap, so a trajectory can contribute to multiple event categories.}
\label{tab:output_counts}
\begin{tabular*}{\linewidth}{@{\extracolsep{\fill}}lr@{}}\toprule
Physical annotation & Trajectories\\\midrule
Target object--finger contact & 61,764\\
Contact formation & 44,859\\
Contact loss & 26,459\\
Object translation during contact & 42,569\\
Object rotation about the vertical axis & 9,491\\
Object tilt & 22,231\\
Grasp establishment & 35,596\\
First grasp maintained until observation ends & 5,313\\\bottomrule
\end{tabular*}
\end{table}

\para{Event criteria}
Contact detection combines target object--finger contact flags with a measured force magnitude above 0.05 N; contact is considered sustained when these conditions hold for three consecutive observations. Motion thresholds adapt to object scale. Let $s$ denote the shorter horizontal extent of the object's collision geometry at its initial orientation. The translation threshold is
\begin{equation}
\delta_x=\operatorname{clip}(0.1s,0.004,0.012)\ \mathrm{m},
\end{equation}
and the angular threshold is $\arcsin(\min(1,2\delta_x/s))$, clamped to $10^\circ$--$25^\circ$. Motion events must exceed the threshold for three consecutive observations and coincide with contact or motion continuing after contact loss. Pushing additionally requires support-state evidence that the object moves along a supporting surface.

Post-grasp slip denotes relative motion between an object and the gripper after the gripper has begun supporting it, with a displacement threshold of $\operatorname{clip}(0.15s,0.006,0.020)$ m. Dropping denotes subsequent separation from the supporting gripper followed by downward motion.

\subsection{Successful Holding and Failed Contact}
Successful holding and failed contact provide complementary interaction experience. Holding captures state persistence after grasp establishment, whereas failed attempts reveal contact-driven object responses. Failed-contact segments are grouped into contracted contact and open contact according to the measured gripper state. The composition experiments use the normalized closure at the final frame of each segment: values above 0.25 indicate contracted contact, and the remaining values indicate open contact. This grouping characterizes the contact configuration within a segment.

\subsection{Reforge-Real: Hardware and Data Capture}
\label{app:real_capture}
\begin{figure*}[!tp]
\centering
\includegraphics[width=\textwidth]{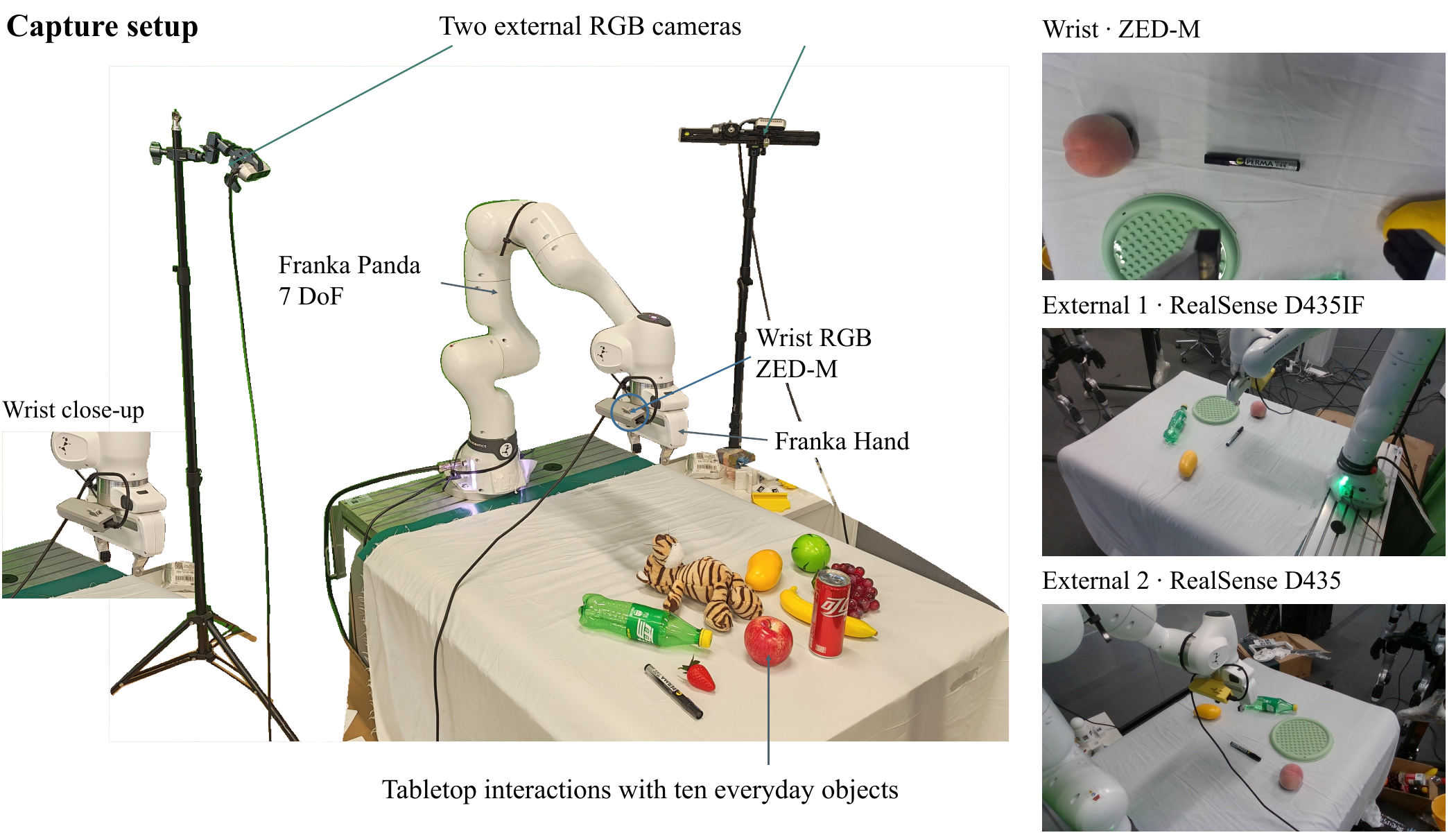}
\caption{Reforge-Real capture setup and three-view observations. The right panel shows the wrist and two external views from the same recorded interaction with a marker.}
\label{fig:real_setup}
\end{figure*}

Reforge-Real uses a seven-degree-of-freedom Franka Panda arm with a Franka Hand electric parallel-jaw gripper, teleoperated through a SpaceMouse and the Deoxys OSC\_POSE interface. \Cref{fig:real_setup} shows the capture setup and synchronized three-view observations; \cref{tab:real_hardware} lists the hardware and acquisition settings.

\begin{table}[!htbp]
\caption{Reforge-Real hardware and acquisition settings. RGB frame rates are requested settings; control and state frequencies are approximate measurements.}
\label{tab:real_hardware}
\centering\tablefont
\begin{tabularx}{\columnwidth}{@{}lY@{}}\toprule
Component & Configuration \\\midrule
Robot arm & Franka Panda, 7 DoF \\
Gripper & Franka Hand, electric parallel-jaw gripper \\
Wrist RGB & ZED-M, rectified left image \\
External RGB 1 & Intel RealSense D435IF \\
External RGB 2 & Intel RealSense D435 \\
Teleoperation & SpaceMouse; Deoxys OSC\_POSE \\
RGB capture & $1280\times720$, requested 15 FPS \\
Control commands & Approximately 20 Hz \\
Arm/gripper states & Approximately 99 Hz/10 Hz \\
Recorded streams & RGB, robot states, control commands, timestamps \\\bottomrule
\end{tabularx}
\end{table}

A wrist-mounted Stereolabs ZED-M and two fixed external cameras, an Intel RealSense D435IF and a D435, observe each interaction. All three RGB streams are configured at $1280\times720$ and 15 FPS. Control commands, end-effector poses, joint and gripper states, and timestamps are recorded alongside the images. End-effector poses are expressed in the robot base frame. Control commands, arm states, and gripper states are recorded at approximately 20, 99, and 10 Hz, respectively.

The collection comprises 30 trajectories for each of 10 everyday object categories, totaling 300 failure-related trajectories. Fruit, beverage containers, plush toys, and markers vary in shape and rigidity, providing diverse contact conditions.

Observation and state streams retain both host and device timestamps. The three views shown in the figures are aligned by nearest-neighbor matching of host monotonic reception timestamps. \Cref{fig:real_sequences} further illustrates the temporal evolution of real-world interactions.
\section{Experimental Settings}
\label{app:setup}
\subsection{Implementation Details}
\label{app:optimization}
\label{app:supervision_settings}
Our experiments cover five backbones from the Cosmos3, DreamDojo, and LingBot-VA families~\cite{cosmos3,dreamdojo,lingbot}. Each model predicts 16 future frames conditioned on the initial observation and a 16-step robot action sequence. Actions comprise end-effector translation, a six-dimensional rotation representation, and gripper state. Each action component is normalized using training-data quantiles as $2(a-q_{.01})/\max(q_{.99}-q_{.01},10^{-8})-1$, where $q_{.01}$ and $q_{.99}$ denote the first and 99th percentiles.

\para{Training and optimization}
We conduct post-training experiments on eight NVIDIA H200 GPUs, using a global batch size of 64, BF16 computation, and FP32 optimizer states. The VAE is frozen, while the generative backbone is updated. We use AdamW with $\epsilon=10^{-8}$ and zero weight decay. The training seed is 62101. Cosmos3 and DreamDojo use $(\beta_1,\beta_2)=(0.9,0.99)$ with a gradient-clipping threshold of 1.0; LingBot-VA uses $(0.9,0.95)$ and 2.0, respectively.

For Cosmos3 and DreamDojo, the learning rate warms up to $10^{-6}$ over 100 steps and then decays linearly. LingBot-VA warms up to $3\times10^{-6}$ over 100 steps and maintains that rate thereafter. Data-composition experiments use the initial portion of the main training schedule. The added robot-action input columns in DreamDojo are initialized to zero and trained. In LingBot-VA, a trainable \texttt{action\_embedder} encodes input actions into features.

\para{Physical supervision}
WorldReforge augments the base generative objective with IA and SC, gradually introducing the auxiliary losses early in training. \Cref{tab:optimization} reports the main-experiment loss weights and feature extraction locations.

IA emphasizes reconstruction in interaction regions, while SC establishes cross-frame correspondences using visible physical surface points. SC performs soft matching directly on backbone features, without an additional projection head. Each segment uses at most 16 object points visible in the initial wrist-view frame, with matching temperature $\tau=0.1$.

\para{IA mask construction}
IA projects the object's eight 3D bounding-box corners into both external views and expands each enclosing rectangle by two pixels on the $640\times544$ canvas. After clipping to view boundaries and mapping to each backbone's latent grid, weights are fractional cell overlaps, summed across views and clipped to $[0,1]$. The margin defines the neighborhood; no separate contact mask, occlusion filtering, or smoothing is used. Masks follow synchronized poses of the same object geometry. Each future latent mask is the pointwise maximum over four source frames. The conditioning mask and IA loss for windows without valid boxes are zero.

\para{SC coordinate and projection conventions}
In main-paper Eq.~(5), $T_t^o$ is object-to-world and $T_t^c$ is optical-camera-to-world; hence $(T_t^c)^{-1}$ is world-to-camera. The homogeneous point $X_i=((\mathbf{x}_i^o)^\top,1)^\top$ is expressed in the simulator object's rigid-body root frame, in meters with mesh scale incorporated. It is obtained by transforming an initially visible ray--surface intersection into this frame and remains fixed as the object moves. Writing each rigid transform in terms of rotation $R$ and translation $\mathbf{t}$ gives
\[
\mathbf{x}_{it}^c=(R_t^c)^\top
\bigl(R_t^o\mathbf{x}_i^o+\mathbf{t}_t^o-\mathbf{t}_t^c\bigr).
\]

\noindent The optical axes are right, down, and forward; the fixed conversion from simulator camera axes is included in $T_t^c$. The intrinsic matrix $K_t$ acts on the three-dimensional camera point, and $\pi(u,v,w)=(u/w,v/w)$ performs perspective division. SC labels use a pinhole projection with no lens-distortion or undistortion step. Native image pixels are then mapped to the training canvas and each backbone's feature grid using the corresponding image-preprocessing transforms. Positive depth, geometric visibility, and valid image/feature support determine which correspondences contribute to SC.

\para{Inference settings}
Inference follows each backbone's sampler and guidance formulation. All three Cosmos3 variants use 30 sampling steps, guidance scale 1, and timestep shift 10. DreamDojo uses 30 steps, guidance scale 0, and timestep shift 5. LingBot-VA uses 5 sampling steps and guidance scale 5. The inference seed is 62100 throughout.

\subsection{Training Data and Experience Recipes}
The main experiments use 25.6k interaction segments for 4,000 post-training steps; failure-related segments account for 45\% of the training data.

Execution-composition experiments examine action coverage, while response-composition experiments examine the complementarity of successful and failed experience. Each execution recipe uses 768 segments for 400 steps, and each response recipe uses 1,536 segments for 600 steps; their compositions are given in \cref{tab:execution_recipes,tab:response_recipes}. The physical-supervision ablation starts from the same pretrained Cosmos3 Nano checkpoint and uses 6.4k segments for 1,000 steps to compare the base objective, IA, SC, joint supervision, and pseudo-label SC.

\begin{table}[!htbp]
\caption{Execution recipes for main-paper Tab.~4. Each recipe contains 768 segments and is trained for 400 steps.}
\label{tab:execution_recipes}
\tablefont\begin{tabular*}{\columnwidth}{@{\extracolsep{\fill}}lrrr@{}}\toprule
Experience & Translation & Rotation & Closure \\\midrule
All types & 256 & 256 & 256 \\
Without translation & 0 & 384 & 384 \\
Without rotation & 384 & 0 & 384 \\
Without closure & 384 & 384 & 0 \\\bottomrule
\end{tabular*}\end{table}

\begin{table}[!htbp]
\caption{Response recipes for main-paper Tab.~5. Each recipe contains 1,536 segments and is trained for 600 steps. Contracted and open refer to measured gripper states during failed contact.}
\label{tab:response_recipes}
\tablefont\begin{tabular*}{\columnwidth}{@{\extracolsep{\fill}}lrrr@{}}\toprule
Experience & Holding & Contracted & Open \\\midrule
Holding & 1,536 & 0 & 0 \\
Holding + contracted & 768 & 768 & 0 \\
Holding + open & 768 & 0 & 768 \\
Mixed & 768 & 384 & 384 \\\bottomrule
\end{tabular*}\end{table}

\begin{table*}[!tp]
\caption{Physical-supervision settings for the main experiments. The listed SC coefficient includes the 0.25 multiplier. Feature grids are given as width $\times$ height, and block indices are zero-based.}
\label{tab:optimization}
\tablefont\begin{tabular*}{\textwidth}{@{\extracolsep{\fill}}lrrrrr@{}}\toprule
Backbone & $\lambda_{\rm IA}$ & $\lambda_{\rm SC}$ & SC block & Feature grid & Channels \\\midrule
Cosmos3 Nano & 0.172320 & 0.339032 & 18 & $20\times17$ & 4,096 \\
DreamDojo & 0.015339 & 0.040554 & 14 & $42\times36$ & 2,048 \\
LingBot-VA & 0.057395 & 0.080585 & 15 & $10\times9$ & 3,072 \\\bottomrule
\end{tabular*}\end{table*}

\subsection{Evaluation Data}
Main-model comparisons and checkpoint analysis use the same 1,500 evaluation videos: 1,200 from Reforge-Eval, 200 from DROID~\cite{droid}, and 100 from Reforge-Real. Reforge-Eval comprises 200 segments for each of six conditions: translation, rotation, closure timing, contracted contact, open contact, and successful holding.

Based on whether an object appears in the post-training data, the simulated evaluation contains 784 segments with seen objects and 416 segments with unseen objects; the latter span 9 objects. Of all simulated segments, 622 carry failure-related annotations. Execution- and response-composition analyses each use the 200 segments in the corresponding category.

\subsection{Motion Evaluation}
Motion evaluation uses a $640\times544$ three-view canvas, with the wrist view above the left and right external views. Query points are determined by geometry visible in the initial frame and shared across models. We use at most ten points each for the object surface and gripper fingers: six in the wrist view and two in each external view. The number of valid points depends on geometry and visibility.

CoTracker3~\cite{cotracker3} tracks queries separately in each generated view. The tracks are mapped to canvas coordinates and compared with reference trajectories obtained by projecting simulator geometry. CoTracker extracts motion from generated videos; SC training supervision in WorldReforge comes from simulator ground truth.

Object and gripper ADE/FDE follow the definitions in the main paper, are measured in pixels, and are averaged with equal weight per video. ADE covers future frames 1--16, while FDE uses frame 16; valid points are determined by ground-truth visibility. The Static baseline holds each query at its initial-frame position.

\subsection{Reconstruction and Visual Quality}
PSNR and SSIM~\cite{ssim} assess future-frame reconstruction over the full three-view canvas. RGB values are normalized to $[0,1]$, and per-video $\operatorname{PSNR}=-10\log_{10}(\operatorname{MSE})$. SSIM uses the color-image implementation in \texttt{skimage}, with data range 1 and the default full-frame settings, averaged over future frames. All metrics are aggregated with equal weight per video; overall results include all evaluation videos across the three domains.

The visual quality pass rate $Q$ is evaluated by Qwen3.8-27B using a fixed binary classification protocol:
\begin{equation}
Q=100\,\frac{N_{\rm normal}}{N_{\rm valid}}.
\end{equation}
Here, $N_{\rm valid}$ counts valid judgments and $N_{\rm normal}$ counts normal judgments. The evaluator receives the initial conditioning frame (F00) and all 16 generated future frames (F01--F16), but no ground-truth future, action labels, or model identity. Temperature is zero, the seed is 62105, and the output limit is 1,400 tokens. A 32-pixel footer below the canvas provides camera-layout and temporal information. Judgments must satisfy the prescribed JSON format and evidence requirements. Missing or invalid judgments are excluded from both the numerator and denominator.

The evaluation considers five types of visual anomaly: entity continuity, object appearance, robot structure, visible motion, and image corruption. A natural failed grasp, drop, or stationary object does not itself constitute an anomaly. Thus, $Q$ measures visible integrity and complements ADE/FDE, which measure motion accuracy. The complete evaluation prompt follows.
\begin{lstlisting}
You are a binary visible-quality inspector of a robotics video. Inspect ALL attached F00-F16 frames, in order. Each image contains the SAME three synchronized camera views: top=wrist, bottom-left=external_left, bottom-right=external_right. The footer is metadata, not scene content. The source is 15 FPS (0.066667 seconds per step), total 1.066667 seconds. Do not judge the sequence at slow publication playback speed.

Decide whether there is a CLEAR, OBVIOUS visible quality defect. normal means no clear defect detected; abnormal means at least one clear defect is visible. Diagnostic defect types: entity_continuity (unexplained disappearance/appearance/teleportation); object_appearance (clear melting, fragmentation, topology/shape breakdown); robot_body (duplicate, detached, disappearing or deformed robot parts); visible_motion (directly visible impossible motion/support, not inferred hidden forces); visual_corruption (severe blur, tearing, corruption, abnormal flicker/camera jumps that materially impair interpretation). Minor texture differences, simulation style, imperfect photorealism, natural motion blur and uncertain details are not enough.

Camera motion, perspective, ordinary occlusion, entering/leaving the field of view, grasp misses, dropped objects, lack of movement, and task failure are NOT by themselves quality defects. RGB cannot prove hidden contact, hidden support or precise 3D penetration. When a reasonable normal explanation remains, do not force an anomaly. A smooth frozen video can be normal for this quality-only metric. Do not infer task success, action following or agreement with a reference. No reference is provided.

Compare early/middle/late frames and synchronized views. F00 is conditioning context; F01-F16 are future observations. Evaluate the entire clip but separately flag whether a clear defect is already present in F00. For a severe single-frame defect, cite that frame and a neighboring frame; a defect need not persist for two frames.

Return ONLY one JSON object with EXACTLY these fields:
{"label":"normal or abnormal","anomaly_types":[],"start_frame":null,"end_frame":null,"conditioning_frame_issue":false,"evidence":[{"frame":0,"view":"wrist or external_left or external_right or cross_view","observation":"direct visible fact"}],"counter_evidence":["reasonable normal explanation checked"],"summary":"short evidence-based decision"}
For normal use empty anomaly_types and null boundaries. For abnormal use one or more diagnostic types listed above and inclusive integer boundaries 0..16. Both labels require evidence from at least two DISTINCT frames, each with a concrete observation and a valid view. conditioning_frame_issue is a boolean, never a causal attribution. Write concise observations, no markdown or extra fields.
\end{lstlisting}
\section{Supplementary Quantitative Analysis}
\label{app:results}
This section addresses three questions: which interaction responses benefit from physical supervision, how simulated experience transfers to real videos, and how these gains vary with backbone scale and training progress.

Throughout the result tables, bold denotes the best result and underlining, where used, the second best within each metric and comparison group.

\subsection{Interaction Conditions}
\begin{table*}[!tp]
\caption{Cosmos3 Nano predictions under six interaction conditions, each containing 200 videos. We compare object motion, gripper motion, and video quality. Tied best values are all highlighted.}
\label{tab:conditions}
\tablefont\begin{tabular*}{\textwidth}{@{\extracolsep{\fill}}llrrrrrrr@{}}\toprule
Condition & Model & O.ADE $\downarrow$ & O.FDE $\downarrow$ & G.ADE $\downarrow$ & G.FDE $\downarrow$ & PSNR $\uparrow$ & SSIM $\uparrow$ & $Q$ $\uparrow$ \\ \midrule
Translation & Pretrained & 7.24 & 12.92 & 6.27 & 6.86 & 22.35 & 0.764 & 93.50 \\
 & Reforge & 4.98 & 8.85 & 2.24 & 3.08 & 26.28 & 0.862 & 99.00 \\
 & WorldReforge & \textbf{4.72} & \textbf{8.42} & \textbf{2.18} & \textbf{2.97} & \textbf{26.49} & \textbf{0.866} & \textbf{99.50} \\
\midrule
Rotation & Pretrained & 8.15 & 14.08 & 5.95 & 7.04 & 21.91 & 0.755 & 96.00 \\
 & Reforge & 6.67 & 10.58 & 2.24 & 3.03 & 25.34 & 0.850 & \textbf{100.00} \\
 & WorldReforge & \textbf{6.51} & \textbf{10.35} & \textbf{2.15} & \textbf{2.86} & \textbf{25.48} & \textbf{0.852} & \textbf{100.00} \\
\midrule
Closure timing & Pretrained & 7.19 & 11.24 & 7.65 & 8.69 & 22.37 & 0.772 & 96.50 \\
 & Reforge & 5.37 & 7.98 & 2.51 & 3.58 & 26.40 & 0.872 & \textbf{99.00} \\
 & WorldReforge & \textbf{5.00} & \textbf{7.47} & \textbf{2.46} & \textbf{3.45} & \textbf{26.59} & \textbf{0.874} & \textbf{99.00} \\
\midrule
Contracted contact & Pretrained & 12.05 & 19.11 & 5.17 & 6.21 & 21.48 & 0.736 & 99.00 \\
 & Reforge & 10.45 & 16.05 & 1.97 & 3.03 & 24.70 & 0.833 & \textbf{99.50} \\
 & WorldReforge & \textbf{10.00} & \textbf{15.21} & \textbf{1.88} & \textbf{2.79} & \textbf{24.85} & \textbf{0.836} & \textbf{99.50} \\
\midrule
Open contact & Pretrained & 7.98 & 13.93 & 3.95 & 4.56 & 22.53 & 0.751 & 99.00 \\
 & Reforge & 5.86 & 10.14 & 1.97 & 2.92 & 26.11 & 0.846 & \textbf{100.00} \\
 & WorldReforge & \textbf{5.59} & \textbf{9.52} & \textbf{1.94} & \textbf{2.86} & \textbf{26.27} & \textbf{0.852} & \textbf{100.00} \\
\midrule
Successful holding & Pretrained & 11.79 & 16.59 & 6.24 & 6.21 & 21.32 & 0.770 & 83.00 \\
 & Reforge & 7.54 & 9.85 & 2.15 & 2.55 & 25.24 & 0.862 & \textbf{100.00} \\
 & WorldReforge & \textbf{7.23} & \textbf{9.31} & \textbf{2.08} & \textbf{2.47} & \textbf{25.42} & \textbf{0.865} & \textbf{100.00} \\
\bottomrule\end{tabular*}
\end{table*}
Condition-wise comparisons clarify the role of physical supervision in execution variations, failed contact, and successful holding. \Cref{tab:conditions} decomposes the main Cosmos3 Nano results into six conditions, evaluating object responses, gripper motion, and video quality separately.

Compared with Reforge, WorldReforge achieves lower object and gripper ADE and higher PSNR in all six conditions, with object-ADE reductions ranging from 2.35\% to 6.96\%. The same physical-supervision objective improves both motion responses during failed contact and state persistence after successful grasping.

\subsection{Successful and Failed Real-World Interactions}
\label{app:droid_outcomes}
\begin{table*}[!tp]
\caption{Video reconstruction and visual quality on successful and failed DROID interactions. Configurations share evaluation videos. Ranks use unrounded values within each backbone and interaction group.}
\label{tab:droid_outcomes}
\tablefont\begin{tabular*}{\textwidth}{@{\extracolsep{\fill}}llrrrrrr@{}}\toprule
Backbone & Model & \multicolumn{3}{c}{Failed interactions} & \multicolumn{3}{c}{Successful interactions} \\
\cmidrule(lr){3-5}\cmidrule(l){6-8}
& & PSNR $\uparrow$ & SSIM $\uparrow$ & Q $\uparrow$ & PSNR $\uparrow$ & SSIM $\uparrow$ & Q $\uparrow$ \\\midrule
DreamDojo & Pretrained & \textbf{17.09} & 0.686 & 89.29 & 17.71 & 0.700 & 85.00 \\
 & Reforge & 16.72 & \textbf{0.689} & \textbf{98.57} & 17.76 & 0.705 & \textbf{96.67} \\
 & WorldReforge & 16.67 & 0.688 & \textbf{98.57} & \textbf{17.88} & \textbf{0.708} & \textbf{96.67} \\
\midrule
LingBot-VA & Pretrained & 16.94 & 0.670 & 68.57 & 17.43 & 0.683 & 66.67 \\
 & Reforge & 17.65 & 0.667 & \textbf{69.29} & 19.65 & 0.713 & 85.00 \\
 & WorldReforge & \textbf{17.81} & \textbf{0.672} & 66.43 & \textbf{19.76} & \textbf{0.716} & \textbf{91.67} \\
\midrule
Cosmos3 Nano & Pretrained & 19.97 & 0.770 & \textbf{100.00} & 21.34 & 0.791 & \textbf{100.00} \\
 & Reforge & 20.79 & 0.790 & 99.29 & 22.56 & 0.810 & \textbf{100.00} \\
 & WorldReforge & \textbf{20.89} & \textbf{0.791} & \textbf{100.00} & \textbf{22.67} & \textbf{0.812} & \textbf{100.00} \\
\bottomrule\end{tabular*}
\end{table*}
Prediction on real interactions reveals the transfer value of simulated experience. \Cref{tab:droid_outcomes} compares reconstruction and visual quality across the three backbone families on DROID's successful and failed interactions.

For Cosmos3 Nano, WorldReforge improves PSNR over Pretrained by 0.92 and 1.34\,dB on the failure and success groups, respectively, with physical supervision further improving PSNR and SSIM over Reforge. LingBot-VA shows the same trend. DreamDojo improves visible integrity, reaching pass rates of 98.57\% and 96.67\% on the two groups. Together, the reconstruction and visual-quality results demonstrate the transfer value of simulated experience: post-training on simulation helps predict both failed attempts and successful operations in the real world.

\subsection{Physical Ground Truth and Visual Correspondences}
\label{app:gt_pseudo}
\begin{table}[!htbp]
\caption{Comparison of physical ground truth and visual pseudo-labels. Cosmos3 Nano is trained on 6.4k segments for 1,000 steps. Pseudo-label SC retains IA and uses CoTracker coordinates and visibility; WorldReforge uses physical ground truth.}
\label{tab:correspondence_sources}
\centering\tablefont\setlength{\tabcolsep}{1.5pt}
\begin{tabular*}{\columnwidth}{@{\extracolsep{\fill}}lrrrrr@{}}\toprule
 & \multicolumn{3}{c}{Reforge-Eval} & DROID & Real \\
\cmidrule(lr){2-4}
Objective & O.ADE$\downarrow$ & G.ADE$\downarrow$ & PSNR$\uparrow$ & PSNR$\uparrow$ & PSNR$\uparrow$ \\
\midrule
Reforge & 7.02 & 2.85 & 24.53 & 21.07 & 21.28 \\
+ IA & 6.84 & \underline{2.84} & \underline{24.58} & 21.08 & \underline{21.30} \\
+ SC & \underline{6.81} & 2.84 & 24.55 & \underline{21.15} & 21.28 \\
Pseudo-label SC & 6.87 & 2.85 & 24.54 & 21.00 & 21.26 \\
WorldReforge & \textbf{6.59} & \textbf{2.66} & \textbf{24.67} & \textbf{21.18} & \textbf{21.30} \\
\bottomrule
\end{tabular*}
\end{table}
Comparing supervision sources tests the value of physical geometry for learning motion across frames. \Cref{tab:correspondence_sources} contrasts physical ground truth with visual pseudo-labels. Pseudo-label SC retains IA and uses CoTracker to estimate future point coordinates and visibility from training RGB videos; initial queries and IA regions are derived from physical ground truth. All configurations start from the same pretrained model and use the same 6.4k segments for 1,000 steps.

Relative to pseudo-label SC, joint supervision with physical ground truth reduces object and gripper ADE by 3.96\% and 6.78\%, respectively. Physical geometry provides a consistent basis for surface-point positions and visibility, yielding cross-frame supervision that improves the accuracy of both object and gripper motion prediction.

\subsection{WorldReforge across Cosmos3 Scales}
\label{app:cosmos_family}
\begin{table*}[!tp]
\caption{Object/gripper motion and simulated/real reconstruction after Cosmos3 Edge post-training.}
\label{tab:cosmos_edge}
\centering\tablefont\setlength{\tabcolsep}{1.5pt}
\begin{tabular*}{\textwidth}{@{\extracolsep{\fill}}lrrrrrrrrrr@{}}\toprule
 & \multicolumn{6}{c}{Reforge-Eval} & \multicolumn{2}{c}{DROID} & \multicolumn{2}{c}{Reforge-Real} \\
\cmidrule(lr){2-7}\cmidrule(lr){8-9}\cmidrule(lr){10-11}
Objective & O.ADE$\downarrow$ & O.FDE$\downarrow$ & G.ADE$\downarrow$ & G.FDE$\downarrow$ & PSNR$\uparrow$ & SSIM$\uparrow$ & PSNR$\uparrow$ & SSIM$\uparrow$ & PSNR$\uparrow$ & SSIM$\uparrow$ \\
\midrule
Pretrained & 8.40 & 13.30 & 7.42 & 9.86 & 21.44 & 0.743 & 19.49 & 0.747 & \textbf{20.47} & 0.847 \\
Reforge & \underline{7.65} & \underline{12.13} & \underline{3.17} & \underline{4.25} & \underline{23.89} & \underline{0.814} & \underline{20.04} & \underline{0.761} & 20.29 & \underline{0.850} \\
WorldReforge & \textbf{7.52} & \textbf{11.99} & \textbf{3.05} & \textbf{4.10} & \textbf{24.00} & \textbf{0.816} & \textbf{20.17} & \textbf{0.763} & \underline{20.39} & \textbf{0.852} \\
\bottomrule
\end{tabular*}
\end{table*}
\begin{table*}[!tp]
\caption{Object/gripper motion and simulated/real reconstruction after Cosmos3 Super post-training.}
\label{tab:cosmos_super}
\centering\tablefont\setlength{\tabcolsep}{1.5pt}
\begin{tabular*}{\textwidth}{@{\extracolsep{\fill}}lrrrrrrrrrr@{}}\toprule
 & \multicolumn{6}{c}{Reforge-Eval} & \multicolumn{2}{c}{DROID} & \multicolumn{2}{c}{Reforge-Real} \\
\cmidrule(lr){2-7}\cmidrule(lr){8-9}\cmidrule(lr){10-11}
Objective & O.ADE$\downarrow$ & O.FDE$\downarrow$ & G.ADE$\downarrow$ & G.FDE$\downarrow$ & PSNR$\uparrow$ & SSIM$\uparrow$ & PSNR$\uparrow$ & SSIM$\uparrow$ & PSNR$\uparrow$ & SSIM$\uparrow$ \\
\midrule
Pretrained & 8.63 & 13.94 & 6.32 & 7.67 & 22.24 & 0.766 & 20.82 & 0.786 & 20.91 & 0.862 \\
Reforge & \underline{7.16} & \underline{11.36} & \underline{2.71} & \underline{3.44} & \underline{24.86} & \underline{0.834} & \underline{21.60} & \underline{0.802} & \underline{21.36} & \underline{0.864} \\
WorldReforge & \textbf{7.06} & \textbf{11.26} & \textbf{2.59} & \textbf{3.33} & \textbf{24.97} & \textbf{0.837} & \textbf{21.70} & \textbf{0.804} & \textbf{21.48} & \textbf{0.867} \\
\bottomrule
\end{tabular*}
\end{table*}
Comparisons on Cosmos3 Edge and Super examine the applicability of WorldReforge across model scales. Both variants start from their respective pretrained checkpoints and use the main 25.6k training segments for 1,000 steps, as reported in \cref{tab:cosmos_edge,tab:cosmos_super}.

Compared with their respective Pretrained models, WorldReforge achieves measured object-ADE reductions of 10.38\% on Cosmos3 Edge and 18.21\% on Super. Gripper prediction also improves, alongside reconstruction on simulation and DROID. These results show that WorldReforge improves object-response and robot-motion prediction at both backbone scales.

\subsection{Training Dynamics}
\label{app:checkpoint_curve}
\begin{figure*}[!tp]
\centering
\includegraphics[width=.86\textwidth]{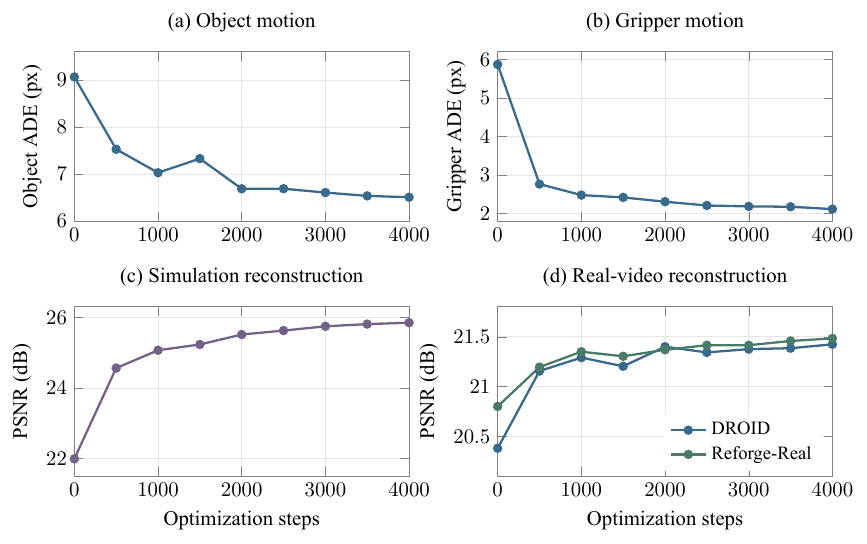}
\caption{Training dynamics of Cosmos3 Nano WorldReforge across nine checkpoints. Motion errors are measured on Reforge-Eval; reconstruction quality is evaluated separately on simulation, DROID, and Reforge-Real.}
\label{fig:checkpoint_curve}
\end{figure*}

\begin{table*}[!tp]
\caption{Cosmos3 Nano WorldReforge across training steps. Step 0 is Pretrained. Motion is evaluated on Reforge-Eval; reconstruction is reported by domain.}
\label{tab:checkpoints}
\tablefont\begin{tabular*}{\textwidth}{@{\extracolsep{\fill}}rrrrrrrrrrr@{}}\toprule
& \multicolumn{6}{c}{Reforge-Eval} & \multicolumn{2}{c}{DROID} & \multicolumn{2}{c}{Reforge-Real} \\
\cmidrule(lr){2-7}\cmidrule(lr){8-9}\cmidrule(l){10-11}
Steps & O.ADE $\downarrow$ & O.FDE $\downarrow$ & G.ADE $\downarrow$ & G.FDE $\downarrow$ & PSNR $\uparrow$ & SSIM $\uparrow$ & PSNR $\uparrow$ & SSIM $\uparrow$ & PSNR $\uparrow$ & SSIM $\uparrow$ \\\midrule
0 & 9.07 & 14.65 & 5.87 & 6.59 & 21.99 & 0.758 & 20.38 & 0.777 & 20.80 & 0.860 \\
1,000 & 7.03 & 10.88 & 2.48 & 3.30 & 25.07 & 0.839 & 21.29 & 0.796 & 21.35 & 0.865 \\
2,000 & 6.69 & 10.32 & 2.31 & 3.15 & 25.52 & 0.850 & \underline{21.40} & \underline{0.797} & 21.37 & 0.866 \\
3,000 & \underline{6.61} & \underline{10.22} & \underline{2.19} & \underline{3.00} & \underline{25.75} & \underline{0.855} & 21.38 & 0.796 & \underline{21.42} & \underline{0.867} \\
4,000 & \textbf{6.51} & \textbf{10.05} & \textbf{2.12} & \textbf{2.90} & \textbf{25.85} & \textbf{0.858} & \textbf{21.42} & \textbf{0.797} & \textbf{21.48} & \textbf{0.867} \\
\bottomrule\end{tabular*}
\end{table*}
Training dynamics characterize WorldReforge's convergence rate and the benefits of continued optimization. \Cref{tab:checkpoints} compares Cosmos3 Nano at different training steps, with step 0 corresponding to Pretrained; \cref{fig:checkpoint_curve} shows the trends across nine checkpoints.

WorldReforge reduces object ADE relative to Pretrained by 22.44\% within the first 1,000 steps, and the reduction reaches 28.19\% at 4,000 steps. The training curves show rapid early gains followed by gradual convergence. Among the nine evaluated checkpoints, the final model achieves the best motion errors and reconstruction metrics in every domain. Post-training therefore improves interaction prediction rapidly, while continued optimization further refines motion details and real-world reconstruction.

\subsection{Object Motion Magnitude and Contact Phase}
\label{app:motion_groups}
Translation and rotation magnitudes are the maximum ground-truth three-dimensional displacement and angular deviation from the initial frame over future frames. We split Reforge-Eval separately into low, medium, and high tertiles of each magnitude.

\begin{table*}[!tp]
\caption{Cosmos3 Nano object ADE by motion magnitude. The 1,200 Reforge-Eval segments are split separately into translation and rotation tertiles.}
\label{tab:motion_magnitude}
\centering\tablefont
\begin{tabular*}{\textwidth}{@{\extracolsep{\fill}}llrrrrr@{}}\toprule
GT motion & Magnitude group & Segments & Static & Pretrained & Reforge & WorldReforge \\\midrule
\multirow{3}{*}{Object translation} & Low: $\leq 4.466\,\mathrm{mm}$ & 400 & 7.73 & 6.23 & \underline{4.68} & \textbf{4.32} \\
 & Medium: $(4.466,\,13.638]\,\mathrm{mm}$ & 400 & 10.04 & 8.92 & \underline{6.81} & \textbf{6.56} \\
 & High: $>13.638\,\mathrm{mm}$ & 400 & 10.37 & 12.05 & \underline{8.94} & \textbf{8.65} \\
\midrule
\multirow{3}{*}{Object rotation} & Low: $\leq 2.756^\circ$ & 400 & 6.97 & 6.45 & \underline{4.65} & \textbf{4.27} \\
 & Medium: $(2.756,\,8.492]^\circ$ & 400 & 8.26 & 8.69 & \underline{6.33} & \textbf{5.99} \\
 & High: $>8.492^\circ$ & 400 & 12.91 & 12.06 & \underline{9.46} & \textbf{9.27} \\
\bottomrule
\end{tabular*}
\end{table*}
In \cref{tab:motion_magnitude}, WorldReforge attains the lowest object ADE in all six magnitude groups. Relative to Reforge, it reduces ADE by 7.80\% in the low-translation group and 8.17\% in the low-rotation group. Physical supervision improves object-motion prediction across translation and rotation magnitudes, including the low-translation and low-rotation groups.

Contact-phase analysis relates these gains to the temporal evolution of an interaction. Measured fingertip contact distinguishes three states: before the first contact, during contact, and after contact when contact is currently absent. Errors are averaged over the relevant frames within each segment, then across segments. WorldReforge reduces object error relative to Reforge by approximately 4\%--5\% in all three states, indicating that improvements extend across contact establishment and subsequent motion.
\section{Paired-Action Responses}
\label{app:paired}
Both branches use the same sampling configuration. Response differences are computed from RGB values normalized to $[0,1]$ over future frames 1--16 of the full three-view canvas. The initial conditioning frame is excluded, and reported RDE values are multiplied by $10^3$.

RDE measures relative responses induced by action changes, while branch-level metrics measure the accuracy of individual predictions. Differencing cancels biases shared by both branches; joint evaluation therefore assesses both relative changes and absolute predictions.
\section{Supplementary Qualitative Results}
\label{app:qualitative}
Multi-view interactions and Cosmos3 Nano predictions are compared with ground truth at matched views and timestamps.

\subsection{Complete Interaction Sequences}

\begin{figure*}[p]
\centering
\includegraphics[width=\textwidth,height=.925\textheight,keepaspectratio]{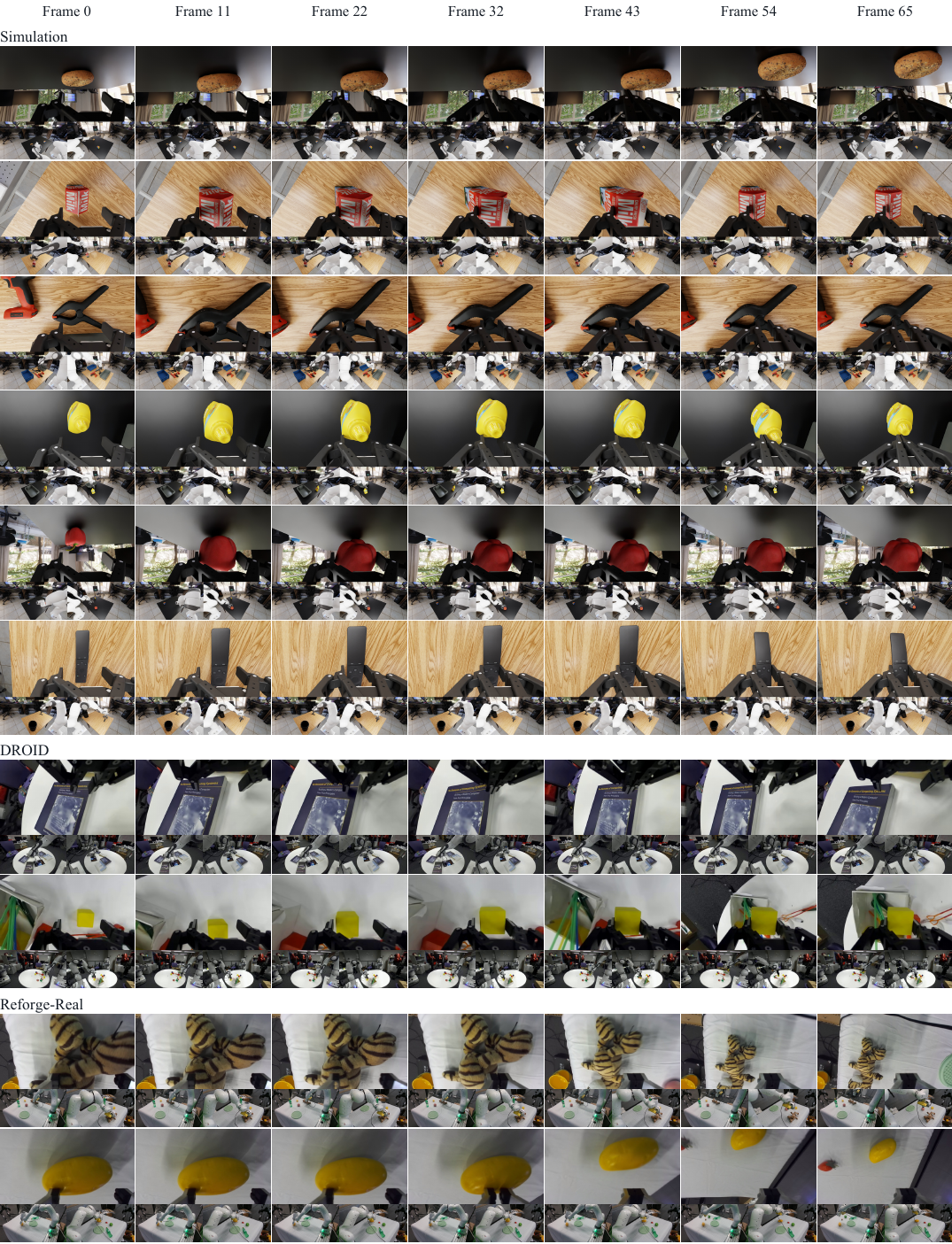}
\caption{Long interaction sequences from Reforge, DROID, and Reforge-Real. Simulated examples show, in order, a bagel, milk carton, spring clamp, mustard bottle, red pepper, and remote control; DROID shows a book and a block; Reforge-Real shows a plush toy and a mango. Each row spans approximately 4.3 seconds at seven timestamps. Each frame retains the full wrist view above the two external views.}
\label{fig:supervision_examples}
\end{figure*}

\Cref{fig:supervision_examples} presents six simulated trajectories and two recordings each from DROID and Reforge-Real. Seven timestamps span the full 66-frame simulated sequences and approximately 4.3 seconds of each real interaction.

\subsection{Three-View Interaction Prediction}
\begin{figure*}[p]
\centering
\includegraphics[width=\textwidth,height=.93\textheight,keepaspectratio]{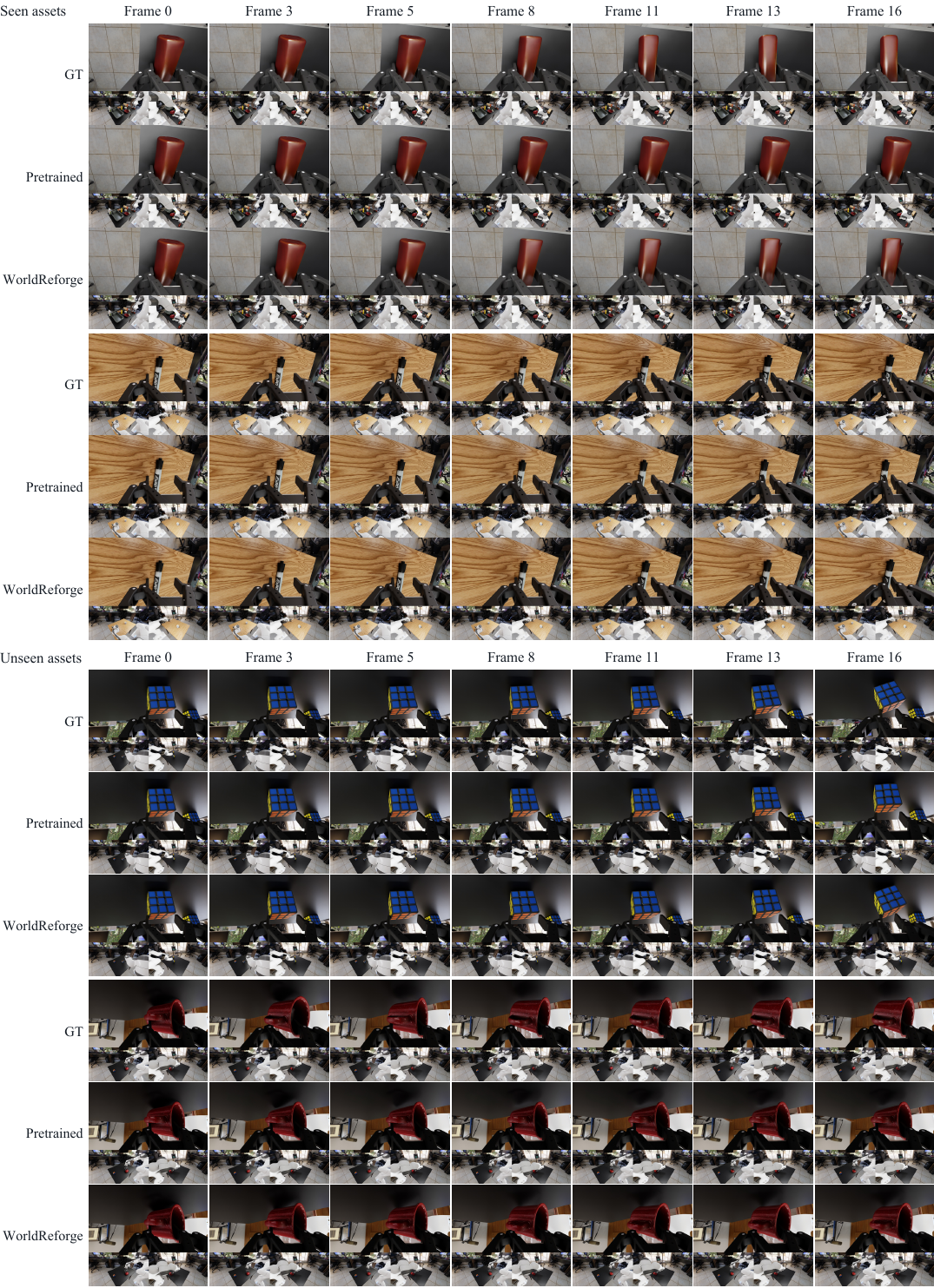}
\caption{Three-view predictions of simulated interactions. The upper two examples show a sauce bottle and marker seen during post-training; the lower two show an unseen Rubik's cube and red mug. Each example presents GT, Pretrained, and WorldReforge, with seven columns spanning frames 0--16. WorldReforge more accurately reproduces the displacement and rotation of these objects.}
\label{fig:prediction_examples}
\end{figure*}

\begin{figure*}[p]
\centering
\includegraphics[width=\textwidth,height=.93\textheight,keepaspectratio]{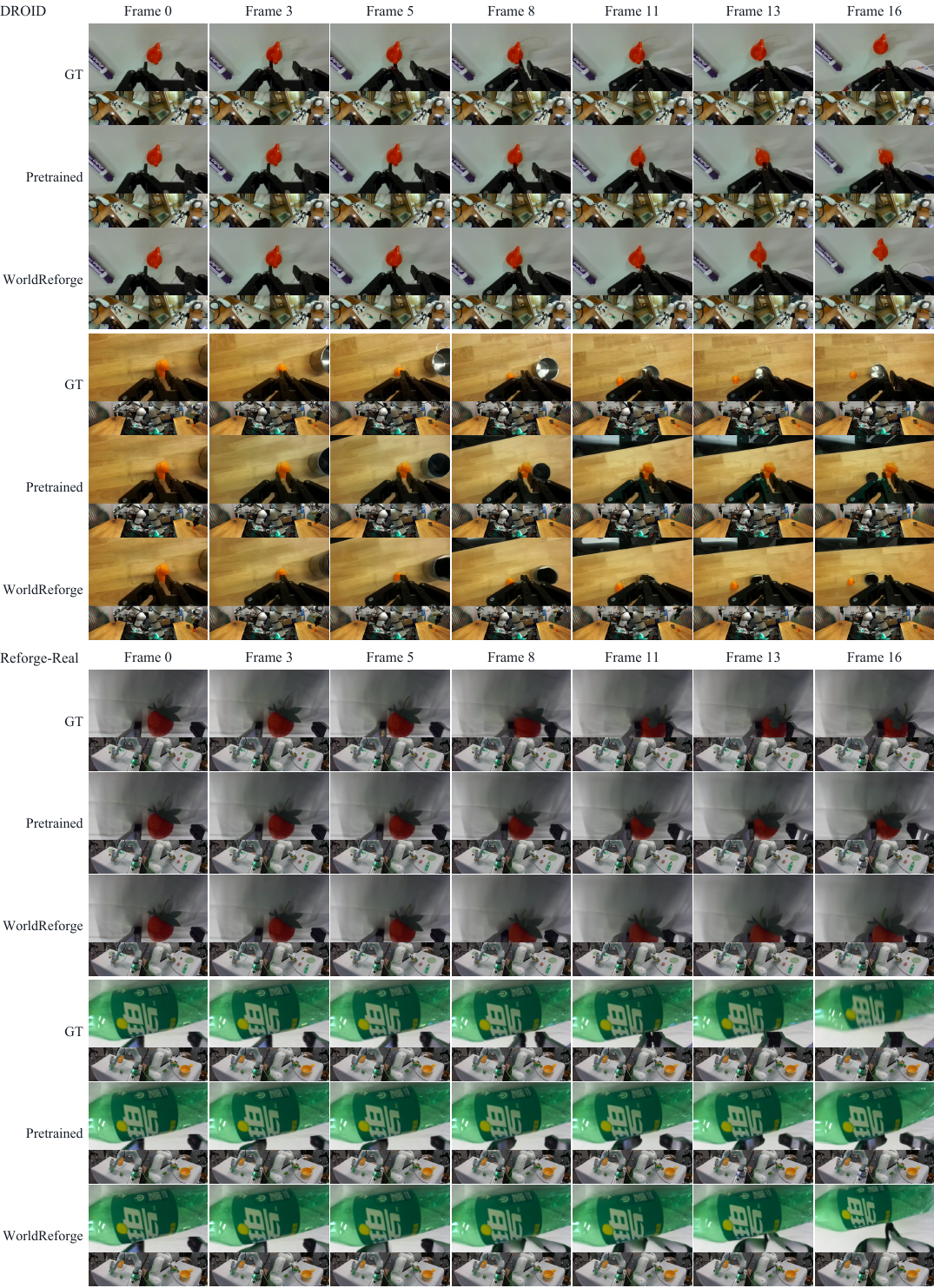}
\caption{Three-view predictions of real-world interactions. The upper two examples show a funnel and orange from DROID; the lower two show contact-induced strawberry displacement and beverage-bottle rotation from Reforge-Real. Each example presents GT, Pretrained, and WorldReforge over seven columns spanning frames 0--16. WorldReforge better follows these objects' changes in position and orientation.}
\label{fig:real_predictions}
\label{fig:real_sequences}
\end{figure*}

\Cref{fig:prediction_examples,fig:real_predictions} compare GT, Pretrained, and WorldReforge on two examples each of seen objects, unseen objects, DROID, and Reforge-Real. Seen objects appear in post-training data.

WorldReforge more accurately predicts contact-driven motion of seen and unseen simulated objects, and better reproduces motion and orientation changes in real videos.

\clearpage
\clearpage
\clearpage

\end{document}